\documentclass{article}

\usepackage[preprint]{neurips_2026}

\usepackage{natbib}
\usepackage[T1]{fontenc}    
\usepackage{hyperref}       
\usepackage{url}            
\usepackage{booktabs}       
\usepackage{amsfonts}       
\usepackage{nicefrac}       
\usepackage{microtype}      
\usepackage{algorithm}
\usepackage{algorithmic}
\usepackage{amsmath}
\usepackage{amssymb}
\usepackage{amsthm}
\usepackage{graphicx}
\usepackage{physics}
\usepackage{multirow}
\usepackage{subfigure}
\usepackage{color}
\usepackage{listings}
\usepackage[most]{tcolorbox}
\usepackage{siunitx}

\lstdefinestyle{promptstyle}{
  basicstyle=\ttfamily\small,
  breaklines=true,
  columns=fullflexible,
  keepspaces=true,
  showstringspaces=false
}

\newtcolorbox{promptbox}[1]{
  colback=gray!3,
  colframe=black!40,
  title=#1,
  fonttitle=\bfseries,
  breakable
}

\DeclareMathOperator{\argmax}{\arg\max}

\newtheorem*{lemma*}{Lemma}

\newtheorem*{theorem*}{Theorem}

\title{Thinking with Patterns: Breaking the Perceptual Bottleneck in Visual Planning via Pattern Induction}

{\author{
  Yichang Jian \\
  State Key Lab of CAD\& CG \\
  Zhejiang University \\
  \texttt{mtdickens1998@gmail.com} \\
  \And
  Boyuan Xiao \\
  State Key Lab of CAD\& CG \\
  Zhejiang University \\
   \texttt{xiaoby202@gmail.com} \\
  \And
  Zhenyuan Huang \\
  State Key Lab of CAD\& CG \\
  Zhejiang University \\
  \texttt{jerryhzy@outlook.com} \\
  \And
  Yifei Peng \\
  State Key Lab of CAD\& CG \\
  Zhejiang University \\
   \texttt{pengyf@zju.edu.cn} \\
  \And
  Yao-Xiang Ding \\
  State Key Lab of CAD\& CG \\
  Zhejiang University \\
  \texttt{dingyx.gm@gmail.com} \\
}}

\begin{document}

\maketitle
\begin{abstract}
Planning from raw visual input remains a significant challenge for current Vision-Language Models (VLMs), when the complexity of input is beyond their one-step perception capability. Motivated by recent advances in Thinking with Images (TWI), a reasonable solution is to decompose the perception process into simpler steps by iteratively acquiring and incorporating local visual evidence. However, even though current VLMs are well-trained in general TWI ability, their perceptual bottleneck in the planning domain remains. To tackle this challenge, we formulate TWI as a tool to gradually build and reflect an accurate internal world model. We find that the resulting training-free planning strategy enables VLMs to solve tasks that are far beyond their initial capabilities, at the cost that too many TWI operations would significantly increase the computational overhead. To further improve efficiency, we propose {\it Pattern Inference}, a novel TWI strategy enabling VLMs to actively recognize known visual patterns in the new tasks and directly infer local world model structures. To obtain these patterns, we propose {\it Pattern Induction}, an online inductive learning strategy treating visual patterns as \textit{composite} and \textit{reusable} experts, which are autonomously discovered and optimized from experience. Experimental evaluations in \textsc{FrozenLake}, \textsc{Crafter} and \textsc{CubeBench} domains show that our approaches achieve a desirable balance between accuracy and efficiency. Project Page: \url{https://future-item.github.io/PI-TWI}.

\end{abstract}
\section{Introduction}
\label{sec:intro}
Due to remarkable reasoning abilities, Vision-Language Models (VLMs) have shown promising potentials in visual planning tasks~\citep{driess2023palmeembodiedmultimodallanguage,mei2024replanvlmreplanningrobotictasks,feng2025reflectiveplanningvisionlanguagemodels,zhao2025cotvlavisualchainofthoughtreasoning}. However, we observe that current VLMs could still struggle in planning domains with raw visual inputs, especially those with complicated relationships among a large number of target objects to be discovered and understood. In our experiments, we show that current VLMs can fail in challenging planning tasks even when provided with ground-truth planners. This suggests a major bottleneck beyond planning ability: The failure essentially lies in the fact that the complexity of input is beyond the capability of VLMs to perceive at once.   

To tackle this perceptual bottleneck, we turn to recent advances in Thinking with Images (TWI)~\citep{openai2025thinkingwithimages,su2025thinking}. This emerging paradigm is characterized by models leveraging visual information as intermediate steps in their thought process by generating intermediate visual cues in the original input, or even predicting purely with visual reasoning traces~\citep{luo2025visual,wu2025dimo,li2025imagine,fu2025refocus,duan2025got,xu2026visual}. The benefit of TWI could include various aspects, while one is particularly related to our problem, such that TWI can be utilized to break the visual perception process of VLMs into simpler steps~\citep{liu2024enhancing}. The key idea lies in iteratively simplifying the visual scenes during reasoning via image modification operations, making the VLMs attend to key aspects related to the current reasoning step. Through this strategy, the one-step perceptual bottleneck is significantly alleviated without directly enhancing the native visual perception ability of VLMs. It is natural to think that the similar idea can also be utilized in planning. Unfortunately, we observe that even though current VLMs are well-trained in general TWI ability, their perceptual obstacle in the planning domain remains (See Table~\ref{tab:pitwi_baseline}). 

To further explore potential solutions, we propose a novel formulation of TWI specially designed for planning tasks. In this formulation, we assume that VLMs are already armed with ground-truth planners in each task domain, functioning only as a tool to gradually build and reflect an accurate internal world model for running the planners. Specifically, we propose a training-free planning strategy that functions as follows: At the beginning, VLM is asked to build a noisy world model from global visual input. During the planning process, VLMs utilize TWI operators to iteratively attend to local parts of the visual input and conducts reflection on the correctness of the local world model structures. We find that given unlimited budget of TWI operations, VLMs are able to solve tasks that are far beyond their initial capabilities. On the other hand, we also recognize a fundamental potential issue: Too many TWI operations would significantly increase the computational overhead, which is measured by token usage in this paper.

To further improve efficiency, we propose {\it Pattern Inference}, a novel TWI strategy centered on the concept of {\it visual patterns}. A visual pattern represents a specific regularity for a local part of the visual input that (1) appears frequently among all inputs from one domain; (2) has a critical impact on the planning results. Assuming that VLMs keep a well-constructed library of such {\it reusable} and {\it composite} patterns, during planning, they can utilize the matching operation to actively recognize known visual patterns in all tasks in the domain and directly infer local world model structures. This operation has much lower cost than the naive structure reflection TWI operation, hence can be utilized to effectively address the previous concern on efficiency. 

To build such pattern libraries, we propose {\it Pattern Induction}, an inductive learning strategy inspired by recent advances in neuro-symbolic inductive learning~\citep{shindo2023alpha,shindo2024learning,piriyakulkij2025poe}. Based on the mixture of experts (MoE)~\citep{jacobs1991adaptive} and self-supervised learning by random masking~\citep{he2022masked} frameworks, the method establishes an online learning strategy that continuously proposes new patterns by VLMs, reweights them using masked past trajectories, and exploits them in pattern inference to improve the efficiency of TWI for planning, allowing visual patterns to be autonomously discovered, optimized, and utilized from experience. 

To summarize, we frame our paper as a proof-of-concept work, targeting at the justification of the following arguments: (1) Formulating TWI as a tool to gradually build and reflect an accurate internal world model could enable VLMs to effectively break their perceptual limitations in planning; (2) Visual patterns could be utilized as reusable and composite tools to significantly reduce the cost of building the internal world model; (3) Critical visual patterns could be autonomously constructed by VLMs based on the inductive online learning framework. We pack our approaches as {\it Pattern-Induced Thinking with Images} (\textsc{PI-TWI}) and evaluate it under three challenging visual planning domains, \textsc{FrozenLake}, \textsc{Crafter}, and \textsc{CubeBench}. The results show that our approaches achieve a desirable balance between accuracy and efficiency, empirically justifying the above arguments.

\section{Related Work}
\label{sec:rw}
\subsection{Planning with VLMs}
VLM-based planning has evolved from grounding language goals in visual observations to increasingly structured forms of long-horizon reasoning. Early embodied multimodal models, such as PaLM-E~\citep{driess2023palmeembodiedmultimodallanguage}, show that visual observations and language instructions can be integrated to support high-level planning for sequential robotic manipulation. Later approaches introduce closed-loop visual feedback, enabling planners such as ReplanVLM~\citep{mei2024replanvlmreplanningrobotictasks} to detect execution failures and revise plans accordingly. More recent methods further move toward world-model-style reasoning: Reflective Planning~\citep{feng2025reflectiveplanningvisionlanguagemodels} imagines future world states to refine long-horizon manipulation plans, while CoT-VLA~\citep{zhao2025cotvlavisualchainofthoughtreasoning} predicts future visual states as intermediate goals before action generation. This line of work has progressively improved the generation, revision, and evaluation of plans. However, this line of work mostly focuses on improving the planning process itself, while assuming that the visual information needed for planning can be reliably extracted from the scene. In contrast, our method targets this perceptual bottleneck by constructing a planner-sufficient world model from local visual evidence and learning reusable visual patterns to impute or prioritize task-critical facts with fewer costly VLM perception calls.

\subsection{Thinking with Images}

Recent advancements have expanded chain-of-thought (CoT) reasoning~\citep{wei2022chain} into the multimodal domain, giving rise to the Thinking of Images (TWI) paradigm. In this broader context, models leverage visual intermediates for a range of cognitive functions. These include invoking visual cropping tools to dynamically acquire evidence \citep{wu2024v, wang2025visuothink,li2025dyfo,zheng2026deepeyes}, utilizing visual scratchpads~\citep{suris2023vipergpt,hu2024visual,su2025openthinkimg,fu2025refocus}, and simulating future states directly in the visual modality ~\citep{li2025imagine,xu2026visual}. Our work builds on this broader view, but focuses on the planning semantics of TWI. In our setting, the use of visual intermediates is to build a symbolic world model that is sufficient for a downstream planner. This requirement introduces structure that is absent from general-purpose TWI: the agent must decide which visual intermediates are relevant to planning and when the current world model is sufficient. Therefore, for our purpose, we formulate TWI as planner-sufficient world model construction.

\subsection{Inductive Learning for Knowledge Acquisition}

Our pattern induction method shares similar spirit with recent efforts in inductive learning for acquiring reusable symbolic knowledge from data. Representative works include differentiable inductive logic programming approaches~\citep{evans2018learning,shindo2023alpha,shindo2024learning}. To induce logical rules from data, these approaches adopt a neuro-symbolic hybrid pipeline, which generates candidate logical clauses from symbolic inference, meanwhile conducts differentiable optimization for learning their fitness weights to the data. Similar methodology is adopted in neural program synthesis methods~\citep{ellis2021dreamcoder,jones2023shapecoder}, where symbolic function libraries are gradually induced and optimized from data, and a neural prediction model is learned to utilize the functions in the library to address the downstream tasks. However, a major challenge lies in how the primitive symbolic knowledge, such as candidate logical clauses and symbolic functions, are generated. Recent efforts have utilized language models for generating the symbolic proposals~\citep{jones2025shapelib,hsu2025programs,piriyakulkij2025poe}. Notably, \citet{piriyakulkij2025poe} considers to utilize programs as composite experts to represent the transition rules in world models. Although we tackle different tasks, the visual patterns in our approach can be treated as a kind of symbolic knowledge to strengthen the visual perception and understanding abilities of VLMs. We hope that our work could inspire further research in the area towards this direction.      

\section{Method}
\label{sec:methxtod}
\begin{figure}[t]
    \centering
    \includegraphics[width=.9\textwidth]{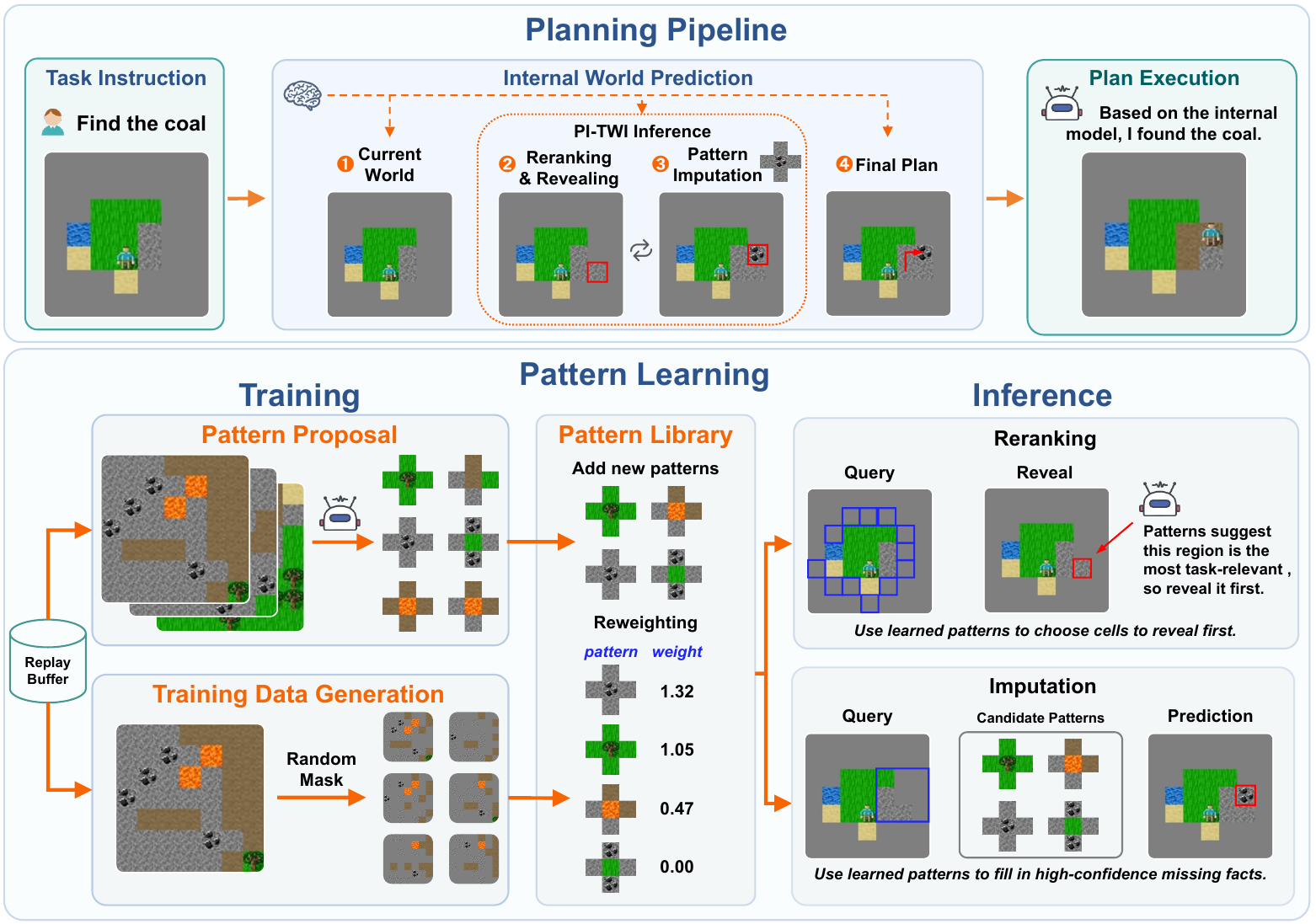}
    \caption{Overview of our \textsc{PI-TWI} approach.} 
    \label{fig:main_pipeline} 
\end{figure}

To break the perceptual limitation of VLMs in visual planning, we formulate Thinking with Images (TWI) as a tool of constructing a planner-sufficient symbolic world model from visual evidence (Sec.~\ref{sec:method_formulation}). For reducing the cost of world model construction, we introduce {\it Pattern Inference}, a novel TWI strategy enabling VLMs to actively recognize known visual patterns in planning tasks and directly infer local world model structures (Sec.~\ref{sec:method_patterns}). To obtain these composite and reusable patterns from learning, we propose {\it Pattern Induction}, an online inductive learning method for building the pattern library from experience (Sec.~\ref{sec:pattern_induction}). We name this framework \textsc{PI-TWI}, short for \emph{Pattern-Induced Thinking with Images}. Fig.~\ref{fig:main_pipeline} provides the overall illustration. 

\subsection{Thinking with Images as Planner-Sufficient World Model Construction}
\label{sec:method_formulation}

We consider a planning instance

\begin{equation}
    x = (I, \mathcal{U}, \{y_u\}_{u \in \mathcal{U}}, \Pi, C, G)
\end{equation}

where $I$ is the input image, $\mathcal{U}$ is a finite set of visual variables, and each $u \in \mathcal{U}$ has a hidden value $y_u$ in a finite domain $\mathcal{Y}_u$. A visual variable is the smallest unit that can be individually inspected by the TWI agent. For example, in \textsc{Crafter}, a visual variable corresponds to a grid cell and its value is a tile or entity type, such as ``grass'', ``water'', or ``stone''. In \textsc{CubeBench}, a visual variable corresponds to a facelet on the cube and its value is one of the six colors, such as ``yellow'' or ``red''. The true values $\{y_u\}_{u \in \mathcal{U}}$ are not available to the agent and must be grounded from the image or inferred from patterns.

A perception operation grounds one selected visual variable through a VLM call. We write

\begin{equation}
    R(I,u) \rightarrow \hat{y}^{R}_{u}
\end{equation}

for the reveal operator, which crops the image region containing $u$ and grounds the observed content to a symbolic value. Each revealed value is a \emph{directly grounded symbolic fact}. A \emph{symbolic fact} is the basic unit of the symbolic world model. At step $t$, the agent maintains two sets of symbolic facts:

\begin{align}
K_t^R &= \{(u,\hat y_u^R): u \text{ has been directly grounded by step } t\}, \\
K_t^I &= \{(u,\hat y_u^I): u \text{ has been imputed by pattern inference by step } t\}.
\end{align}

where $K_t^R$ is the set of \emph{directly grounded symbolic facts} which we have discussed above, and $K_t^I$ is the set of \emph{pattern-imputed symbolic facts} formualted in Eq.~\ref{eq:sym_pattern_inference}.

For a set of facts $K$, let $\operatorname{dom}(K)=\{u:(u,\cdot)\in K\}$. The current symbolic world model is

\begin{equation}
    M_t = K_t^R \oplus K_t^I
    = K_t^R \cup \{(u,\hat y_u^I) \in K_t^I : u \notin \operatorname{dom}(K_t^R)\},
\end{equation}

where $\oplus$ basically means that \textbf{directly grounded facts override pattern-imputed facts whenever both are available for the same variable}, since per-variable perception is accurate, yet imputation is not.

The downstream planner for this task is denoted by $\Pi$. Since the planner may fail or return uncertified plans when the world model is incomplete, we assume an accompanying planner-sufficiency checker

\begin{equation}
    C(M_t) \in \{0,1\}.
\end{equation}

The checker returns $1$ when $M_t$ contains enough information for $\Pi$ to produce a plan satisfying the task criterion. In feasibility-oriented domains such as \textsc{Crafter} and \textsc{CubeBench}, sufficiency means that the current model supports a feasible plan. In optimality-oriented domains such as shortest-path planning in \textsc{FrozenLake}, sufficiency additionally requires that the returned plan can be certified as optimal regardless of the remaining unknown facts.

Our goal is to construct a sufficient and (largely) accurate world model with as few costly perception operations as possible. For each domain, we use a \emph{policy-generation procedure} $G$ as a grounding controller. Given the current world model $M_t$, $G$ generates a set of candidate visual variables that are most worthy of being inspected next:

\begin{equation}
    G(M_t) \subseteq \mathcal{U} \setminus \operatorname{dom}(M_t)\quad \operatorname{dom}(M_t)=\{u:(u,\cdot)\in M_t\}
\end{equation}

where $\operatorname{dom}(M_t)$ is basically the set of all visual variables in $M_t$.

Intuitively, $G$ does not itself solve the planning task as a learned policy. Instead, it generates the next candidates that the agent might inspect in order to make the world model sufficient for planning. Given a rendered raw visual planning instance, an agent may actively inspect parts of the image, ground the inspected content into symbolic facts, and pass the resulting world model to a downstream planner. Since we focus on the perceptual part, we only consider the construction of the world model, and simply assume that this downstream planner is a ground-truth planner (Details are introduced in Appendix~\ref{app:planner}), in the sense that given a correct world model, the planner would produce a correct solution. In our experiments, $G$ is instantiated by LazySP~\citep{dellin2016unifying} for \textsc{FrozenLake}, by a task-and-motion-planning-style search procedure~\citep{garrett2021integrated} for \textsc{Crafter}, and by random selection for \textsc{CubeBench}. These procedures are deliberately task-specific: they reveal only variables that are relevant to the planner and associated checker, rather than asking the VLM to transcribe all variables at once.

Using $G$ without pattern inference gives the ablation of our method without pattern inference, which we denote as \emph{\textsc{PI-TWI} w/o inference}. This ablation repeatedly samples a variable $u_t \in G(M_t)$, reveals it through $R(I,u_t)$, updates $M_t$, and stops once $C(M_t)=1$. It exploits the structure of each planning task, but it treats every instance independently and therefore cannot reuse patterns observed and induced from previous instances. This should be distinguished from \emph{native TWI method}, such as OpenAI-style TWI~\citep{openai2025thinkingwithimages}, which choose their own tool-use and visual-inspection strategy in a largely task-agnostic manner. \textsc{PI-TWI} builds on the task-specific policy-generation procedure by learning a pattern library online and using it to impute or prioritize visual variables before spending additional costly VLM perception calls.

\subsection{Pattern Inference as Gated Mixture-of-Experts}
\label{sec:method_patterns}

A symbolic pattern $\rho$ is a \textit{composite}, \textit{reusable}, \emph{conditional} regularity that can be used to predict the value of a target visual variable. We represent the pattern library as $\mathcal{L}=\{\rho_i\}_{i=1}^m$. Each pattern $\rho_i$ consists of three components: (i) an applicability predicate $a_i(u)\in\{0,1\}$ indicating whether the pattern can be used for target variable $u$; (ii) a \emph{gate} $g_i(u,M_t)\in\{0,1\}$ indicating whether the required contextual facts of the pattern are present and matched in $M_t$; and (iii) a prediction $\mu_i(u)\in\mathcal{Y}_u$ for the value of $u$ when the gate is satisfied. This formulation covers both spatially local patterns, such as repeated tiles in a grid, and more structured patterns, such as facelet-color constraints of a cube.

We view each pattern as a gated expert. For the target variable, if a pattern is applicable and its gate is satisfied, it contributes a smoothed categorical prediction:

\begin{equation}
q_i(v\mid u,M_t)=
\begin{cases}
1-\varepsilon, & v=\mu_i(u),\\
\frac{\varepsilon}{|\mathcal{Y}_u|-1}, & v\neq\mu_i(u)
\end{cases}
\label{eq:pattern_expert}
\end{equation}

where $\varepsilon$ is a small smoothing constant, ensuring non-zero probability. Let $w_i$ be the non-negative reliability weight of pattern $i$. The set of active experts for target variable $u$ is

\begin{equation}
    \mathcal{A}(u,M_t)=\{i: a_i(u)g_i(u,M_t)=1\}.
\end{equation}

If at least one pattern is active, the pattern library predicts the weighted average of all active ones:

\begin{equation}
    p_{\boldsymbol w}(v\mid M_t)
    = \frac{\sum_{i\in\mathcal{A}(u,M_t)} w_i q_i(v\mid u,M_t)}
    {\sum_{i\in\mathcal{A}(u,M_t)} w_i}
\label{eq:sym_pattern_inference}
\end{equation}

If no pattern is active, the pattern library will fall back to the uniform distribution over $\mathcal{Y}_u$. For each $v$, $p_{\boldsymbol w}(v\mid M_t) = \frac{1}{|\mathcal{Y}_u|}$.

In conclusion, (i) only active experts influence the prediction, and (ii) the learned weights determine how much each active pattern should be trusted, relatively.

\paragraph{Pattern-based imputation.}

Given $M_t$, \textsc{PI-TWI} can add value prediction with high confidence to the world model $M_t$, without directly revealing them. For an unrevealed variable $u$, we define
\begin{equation}
    \hat y^I_u = \argmax_{v\in\mathcal{Y}_u} p_{\boldsymbol w}(v\mid M_t),
    \quad
    c_u = \max_{v\in\mathcal{Y}_u} p_{\boldsymbol w}(v\mid M_t).
\end{equation}

where $\hat y^I_u$ is the predicted value of $u$, and $c_u$ is the confidence. If $c_u \geq \tau$, where $\tau$ is a confidence threshold, we add $(u,\hat y^I_u)$ to the set of pattern-imputed symbolic facts $K_t^I$. Since imputed facts no longer need to be perceived, they can help reduce the number of VLM perception calls.

\paragraph{Pattern-based reranking.}

Patterns can also help decide which variable to reveal next. When the policy-generation procedure $G(M_t)$ returns multiple candidates, instead of revealing a randomly sampled one, PI-TWI assigns the candidates a task-dependent score derived from $p_{\boldsymbol w}(y_u\mid M_t)$ and reveals the highest-scoring one. For example, in \textsc{Crafter}, when the current task is to find iron, we rank candidates by $p_{\boldsymbol w}(y_u=\text{iron}\mid M_t)$. Reranking doesn't impute unrevealed variables, but spends perception calls on variables that are more likely to help the planner.

\subsection{Online Inductive Learning for Building Pattern Library}
\label{sec:pattern_induction}
\textsc{PI-TWI} learns its pattern library online from previous planning instances. First, we start with an empty pattern library $\mathcal{L}$ and replay buffer $\mathcal{B}$. The replay buffer $\mathcal{B}$ contains final symbolic world models from past instances. During the process, the agent stores the final symbolic world model of each instance into the replay buffer, propose patterns, adjust their weights, and then use the pattern library for probability inference. As more instances are solved, the buffer provides increasingly informative evidence for pattern learning.

\paragraph{Pattern proposal by VLM}
We use a VLM as a flexible pattern proposer. Given a batch of symbolic world models sampled from $\mathcal{B}$, the VLM gets the batch as input, and propose feasible candidate \emph{macro symbolic patterns}, which will be then parsed into multiple symbolic , deduplicated, and added to $\mathcal{L}$. Each valid macro symbolic pattern will be parsed into one or more symbolic patterns.

\paragraph{Training data generation by masking}

Inspired by previous work ~\citep{he2022masked}, we construct training data for self-supervised learning by randomly masking symbolic facts from world models in $\mathcal B$. For each final world model $M^{(b)}\in\mathcal{B}$, when generating its $r$-th mask, we randomly mask a subset of it $H^{(b,r)}$ to obtain a partially observed model $\tilde{M}^{(b,r)}$ , where $\tilde {M}^{(b,r)} = M^{(b)} \setminus  H^{(b,r)}$. This procedure will finally result in a dataset $\mathcal{D}=\{(\tilde M^{(b,r)}, H^{(b,r)})\}$.

\paragraph{Pattern reweighting}

Once we have the proposed patterns added and training data $\mathcal{D}$ generated, we can infer the probability and perform maximum likelihood estimation to obtain the weights $\boldsymbol{w}$:

\begin{equation}
    \boldsymbol w^* = \argmax_{\boldsymbol w}
    \sum_{(\tilde M^{(b,r)}, H^{(b,r)})\in\mathcal{D}}
    \left[
        \sum_{(u,y_u) \in H^{(b,r)}}
        \log p_{\boldsymbol w}(y_u\mid \tilde  M^{(b,r)}).
    \right]
\label{eq:pattern_reweighting}
\end{equation}

During learning, only the scalar pattern weights $\boldsymbol{w}$ are optimized, and the patterns proposed by the VLM are kept fixed until the next proposal round. Since the number of active patterns is relatively small in our tasks, we use L-BFGS for quicker optimization. The proposal, masking, and reweighting steps are repeated periodically during the online learning process, allowing \textsc{PI-TWI} to improve as the procedure goes on.

\section{Experiment}
\label{experiments}
\begin{table}[t]
\centering
\footnotesize
\setlength{\tabcolsep}{4pt}
\caption{Accuracy comparisons over baselines. G. Acc. and P. Acc. are short for grounding and planning accuracy respectively, both reported in percent. Three representative base models, GPT-5.4~\citep{openai2026gpt54}, Gemini 3.1 Pro~\citep{google2026gemini31pro}, and Qwen3 VL 235B A22B~\citep{bai2025qwen3vltechnicalreport} are introduced.}
\begin{tabular}{llcccccc}
\toprule
\multirow{2}{*}{Environment} & \multirow{2}{*}{Model}
& \multicolumn{2}{c}{VLM direct output}
& \multicolumn{2}{c}{Native TWI}
& \multicolumn{2}{c}{PI-TWI} \\
\cmidrule(lr){3-4}\cmidrule(lr){5-6}\cmidrule(lr){7-8}
& & G. Acc. & P. Acc. & G. Acc. & P. Acc. & G. Acc. & P. Acc. \\
\midrule

 & GPT-5.4
& 92.88 & 85.00 & 92.46 & 85.00 & \textbf{97.19} & \textbf{90.78} \\
\textsc{FrozenLake} & Gemini 3.1 Pro
& 93.87 & 87.50 & 93.66 & 85.00 & \textbf{97.15} & \textbf{94.00} \\
 & Qwen3 VL 235B A22B
& 54.92 & 17.50 & 45.78 & 17.50 & \textbf{95.35} & \textbf{82.78} \\

\midrule

 & GPT-5.4
& 46.39 & 0.00 & 44.28 & 0.00 & \textbf{100.00} & \textbf{100.00} \\
\textsc{Crafter} & Gemini 3.1 Pro
& 26.71& 0.00& 48.78& 41.67& \textbf{100.00} & \textbf{100.00} \\
 & Qwen3 VL 235B A22B
& 0.00 & 0.00 & 22.08 & 0.00 & \textbf{100.00} & \textbf{100.00} \\

\midrule

 & GPT-5.4
& 25.73 & 0.00 & 98.83 & 96.00 & \textbf{100.00} & \textbf{100.00} \\
\textsc{CubeBench} & Gemini 3.1 Pro
& 18.65 & 0.00 & 91.44 & 71.00 & \textbf{100.00} & \textbf{100.00} \\
 & Qwen3 VL 235B A22B
& 16.00 & 0.00 & 17.08 & 0.00 & \textbf{100.00} & \textbf{100.00} \\

\bottomrule
\end{tabular}
\label{tab:pitwi_baseline}
\end{table}

We conduct experiments to justify the following points: (1) Does VLM perception bottleneck exist in challenging visual planning tasks, even with VLM's native TWI abilities? (2) Does our formulation of TWI as a tool for world model construction effectively address the perceptual bottleneck? (3) Can our Pattern Inference method effectively reduce the cost of TWI operations? (4) Does our Pattern Induction method build effective and generalizable pattern library through online inductive learning?

\paragraph{Benchmarks.}

Our experiments focus on three visual planning domains with different structures and objectives: shortest-path planning in \textsc{FrozenLake}~\citep{towers2024gymnasium}, long-horizon resource acquisition and crafting in \textsc{Crafter} environment~\citep{hafner2021benchmarking}, and Rubik’s Cube planning in \textsc{CubeBench}~\citep{gao2025cubebench}. Details of the benchmarks are included in Appendix~\ref{app:bench}.

\paragraph{Baselines and Ablations.}

We compare \textsc{PI-TWI} against the following baselines and ablations: (1) \textbf{VLM direct output} asks the VLM to directly transcribe the rendered image into a symbolic world model in a single pass. This baseline tests whether the model can recover the full symbolic state without active visual grounding. (2) \textbf{Native TWI} lets an off-the-shelf agentic TWI system decide how to inspect or reason over the image using its own tool-use loop, without our task-specific policy-generation procedure $G$, planner-sufficiency checker $C$, replay buffer $\mathcal{B}$, or pattern library $\mathcal{L}$. In our experiments, this corresponds to native systems such as GPT-5.4 with Code Interpreter and Qwen Agent. This baseline is largely task-agnostic: it actively uses visual tools, but it is not given the task-specific $G$ used by \textsc{PI-TWI}. (3) \textbf{\textsc{PI-TWI} w/o inference} uses the same reveal operator $R$ and task-specific policy-generation procedure $G$ as \textsc{PI-TWI}, but disables pattern-based imputation and reranking. The planner can only use directly revealed facts. Since it also does not call a VLM for pattern proposal, its proposal-token cost is zero. (4) \textbf{\textsc{PI-TWI} w/o reweighting} uses the same pattern library as \textsc{PI-TWI}, but disables online optimization of the pattern weights. This tests whether dynamically reweighting patterns using masked trajectories from replay buffer is necessary.

\paragraph{Evaluation Metrics.}

To evaluate the performance of PI-TWI in planning tasks, we focus on both efficiency and accuracy: (1) \textbf{Efficiency.} We employ total token consumption as a proxy for efficiency, comprising \textit{perception} and \textit{proposal} tokens. Total and perception tokens are averaged over all episodes, while proposal tokens are averaged over all proposals. (2) \textbf{Accuracy.} We assess performance using two primary metrics: {\it planning accuracy} and {\it grounding accuracy}. Planning accuracy evaluates the ability to generate a sequence of actions that successfully finish the goal, measured as the ratio of successful episodes to total episodes. Grounding accuracy evaluates the ability to correctly represent the state space. It is defined as the ratio of correctly identified grids over the total number of grids:
\begin{equation}
\text{Grounding Accuracy} = \frac{N_{\text{correct\_imp}} + N_{\text{correct\_per}}}{N_{\text{imp}} + N_{\text{per}}}
\end{equation}
where $N_{\text{imp}}$ and $N_{\text{per}}$ denote the total number of imputed and perceived visual variables, and $N_{\text{correct\_imp}}$ and $N_{\text{correct\_per}}$ represent the number of correctly imputed and perceived visual variables.

In summary, our evaluation covers five dimensions: perception/proposal/total tokens, and planning/grounding accuracy. Note that we also use {\bf reveal count} in our analysis, which only serves as a diagnostic measure to explain the source of token reduction, since under a fixed reveal operator, perception-token cost scales linearly with the number of perception calls; the main efficiency comparison is still based on total token consumption. We elaborate this matter further at Appendix~\ref{sec:token_reveal_count_relation}. Also, unless otherwise specified, quantitative comparisons in tables and experiment section are based on the sum of total input and output tokens.

\begin{table}[t]
\centering
\footnotesize
\setlength{\tabcolsep}{3.5pt}
\caption{Efficiency comparisons using GPT-5.4~\citep{openai2026gpt54}. Accuracy is reported in percent; token costs are reported as input/output token counts, with $k$ denoting thousands. }
\begin{tabular}{llccccc}
\toprule
Environment & Method & Grounding Acc $\uparrow$& Planning Acc $\uparrow$& Prop. token $\downarrow$& Perc. token $\downarrow$& Total token $\downarrow$\\
\midrule
\multirow{4}{*}{FrozenLake} & PI-TWI & 97.19 & 90.78 & \textbf{463.7/267.8} & \textbf{3.10k/158.1} & \textbf{3.19k/208.9} \\
 & w/o inference& \textbf{100.00} & \textbf{100.00} & -- & 5.46k/278.4 & 5.46k/278.4 \\
 & w/o reweight& \textbf{100.00} & \textbf{100.00} & 471.1/268.2 & 5.46k/278.4 & 5.55k/329.4 \\
 
\midrule
\multirow{4}{*}{Crafter} & PI-TWI & \textbf{100.00} & \textbf{100.00} & \textbf{2.73k/208.8} & \textbf{141.9k/3.86k} & \textbf{142.4k/3.90k} \\
 & w/o inference& \textbf{100.00} & \textbf{100.00} & -- & 224.5k/6.10k & 224.5k/6.10k \\
 & w/o reweight& \textbf{100.00} & \textbf{100.00} & \textbf{2.73k/208.8} & 180.2k/4.90k & 180.8k/4.94k \\

\midrule
\multirow{4}{*}{CubeBench} & PI-TWI & \textbf{100.00} & \textbf{100.00} & \textbf{1.56k/499.3} & \textbf{4.31k/244.9} & \textbf{4.45k/289.9} \\
 & w/o inference& \textbf{100.00} & \textbf{100.00} & -- & 4.75k/270.0 & 4.75k/270.0 \\
 & w/o reweight& \textbf{100.00} & \textbf{100.00} & 1.56k/500.1 & 4.75k/270.0 & 4.89k/315.0 \\

\bottomrule

\end{tabular}

\label{tab:pitwi_ablation}
\end{table}

\begin{figure}[t]
    \centering
    \subfigure[]
    {
        \includegraphics[width=0.3\textwidth]{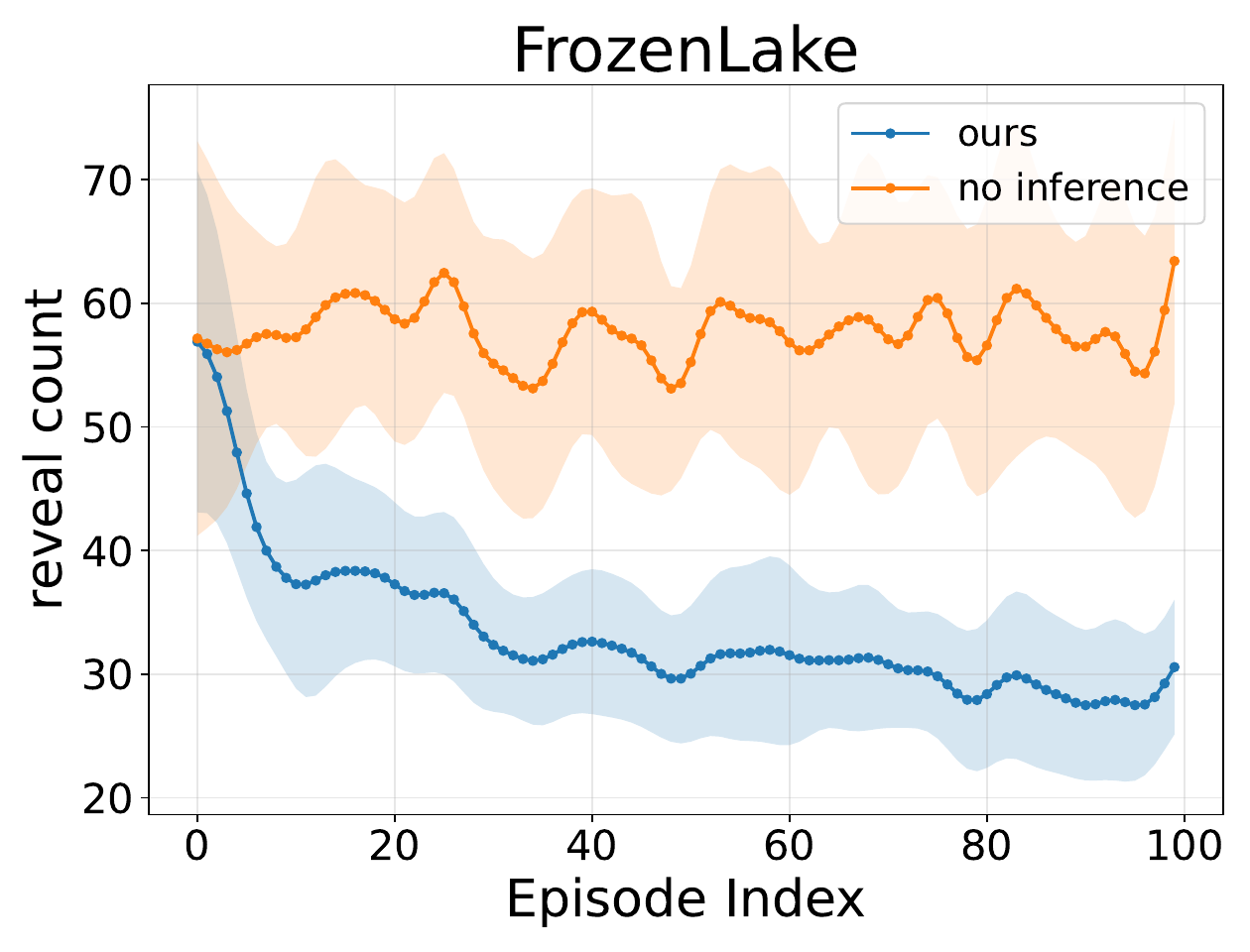}
    }
    \subfigure[]
    {
        \includegraphics[width=0.3\textwidth]{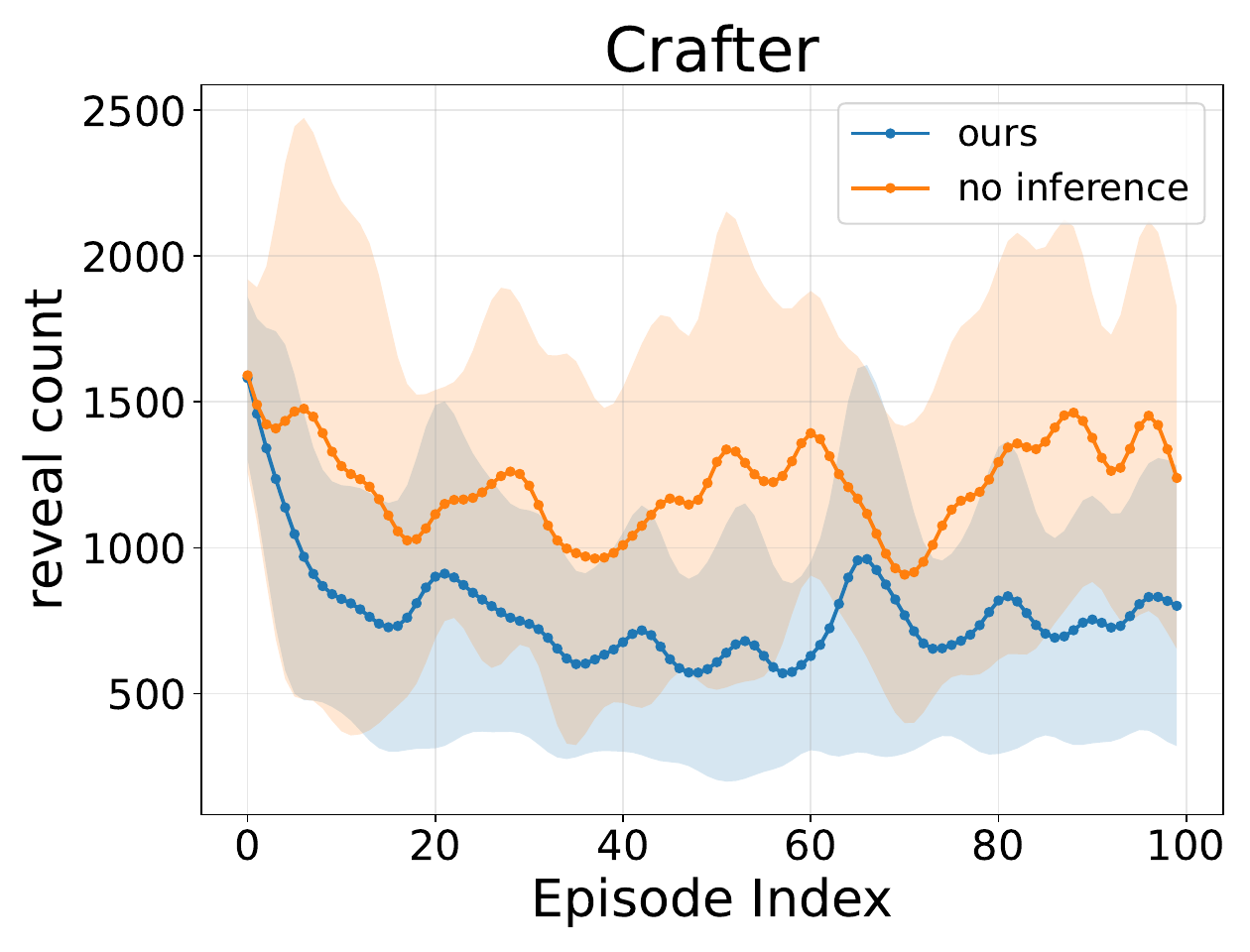}
    }
    \subfigure[]
    {
        \includegraphics[width=0.3\textwidth]{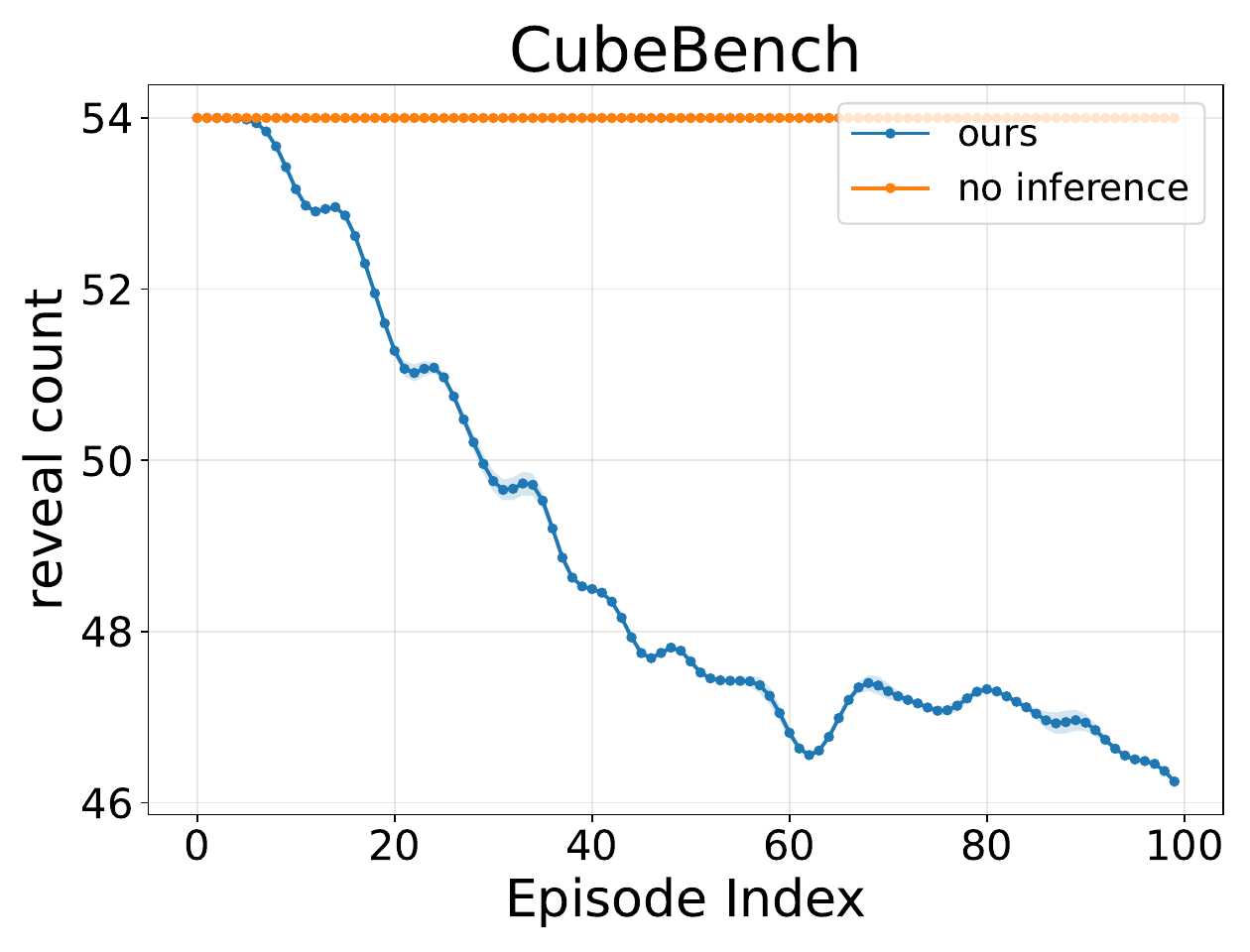}
        \label{subfig:cube_sigma}
    }

    \caption{Reveal count comparisons in the efficiency experiment (gaussian smoothing, $\sigma=2.0$).}
    \label{fig:reveal_count_charts}
\end{figure}

\subsection{Accuracy Comparison}

As shown in Table~\ref{tab:pitwi_baseline}, directly using VLM outputs leads to bad performance, illustrated by zero planning accuracy in visually complex or large-scale environments, like \textsc{CubeBench} and \textsc{Crafter}. Native TWI improves performance for some environment--model pairs, but mainly on visually complex small-scale tasks like \textsc{CubeBench}; however, its effectiveness is inconsistent across models and does not scale reliably to large-scale environments such as \textsc{Crafter}. This suggests that current TWI frameworks and the agentic behavior of foundation models may still be insufficient for these problems. In contrast, we formulate TWI-guided planning as a process of incrementally constructing a symbolic world model. Our full method, \textsc{PI-TWI}, is built upon this formulation. As shown in Table~\ref{tab:pitwi_baseline}, \textsc{PI-TWI} consistently improves both grounding and planning accuracy across all evaluated tasks and models.

\subsection{Efficiency Comparison}

\begin{figure}[t]
    \centering
    \includegraphics[width=.8\textwidth]{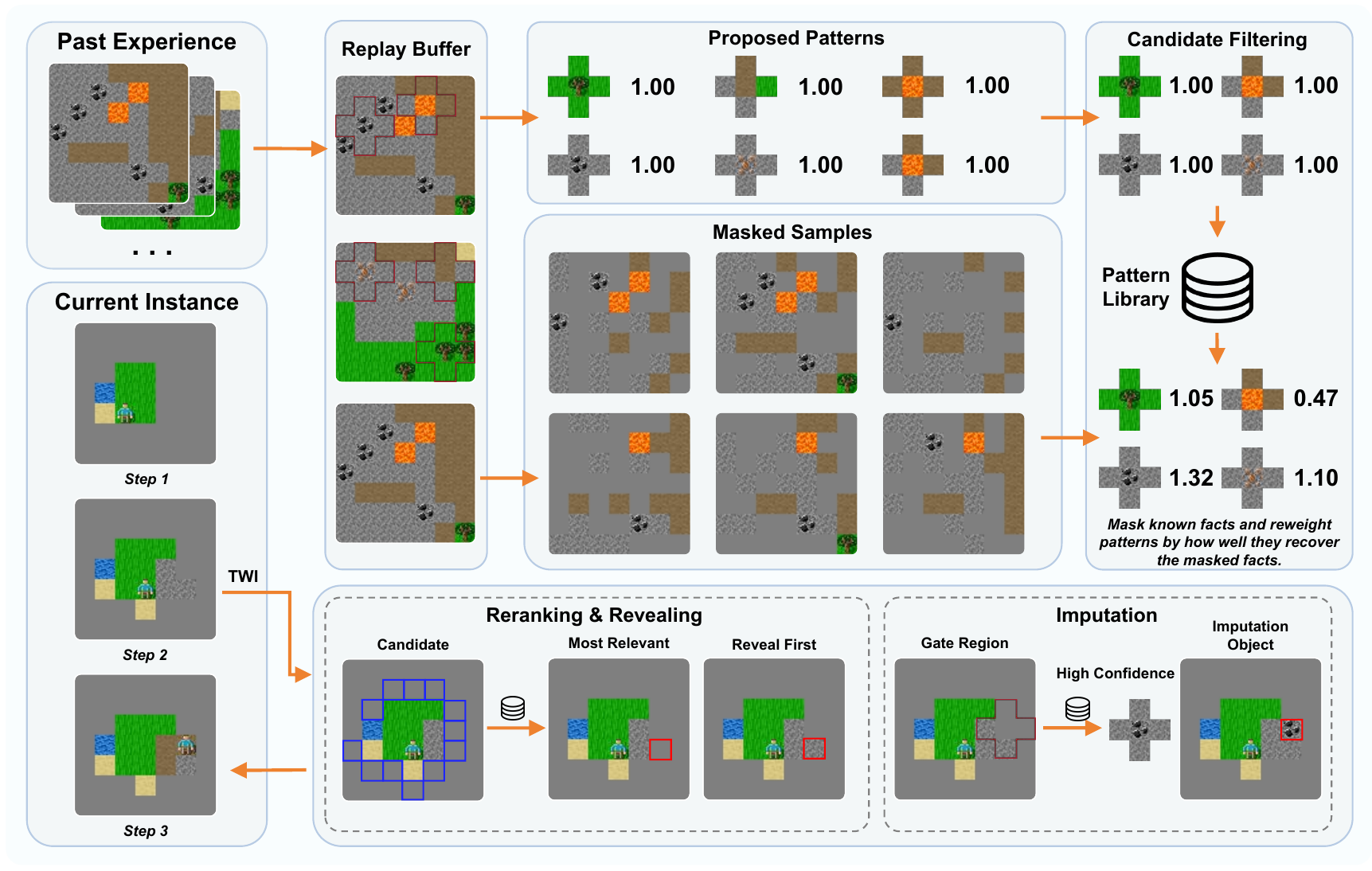}
    \caption{Qualitative illustration of the pattern inference and induction processes in \textsc{Crafter}.} 
    \label{fig:qualitative_craft} 
\end{figure}
As illustrated in Table~\ref{tab:pitwi_ablation}, our method PI-TWI significantly outperforms alternative settings across all tasks in terms of total token consumption. In \textsc{FrozenLake}, although the planning accuracy with PI-TWI decreases to 90.78\%, we argue that this represents a favorable trade-off between efficiency and accuracy. Specifically, we achieve a 40.77\% reduction in total token consumption compared to the ``w/o inference'' baseline, at the expense of less than 10\% decline in accuracy. Similarly, in \textsc{Crafter}, with the help of reranking, the total token consumption is also reduced to 63.44\% of the ``w/o inference'' ablation. In \textsc{CubeBench}, while the total token consumption only drops from 5.02k to 4.74k\footnote{$4.75k+270=5.02k, 4.45k+289.9=4.74k$}, this reduction is significant when considering the theoretical limit. Given that the theoretical minimum of reveal count is 46 (accounting for the 8 corner patterns) and associated minimum of 4.05k input perception tokens, our method achieves around 42.86\% of the maximum possible reduction in terms of input perception tokens~\footnote{$\frac{4.75-4.45}{4.75-4.05} = 42.86\%$. This is calculated using \emph{input perception} tokens, instead of the sum of total input and output tokens.}, which constitutes a substantial improvement in relative terms. The result shows that the reweighting mechanism is also crucial. Since VLM-proposed patterns are not always correct or equally useful, treating all patterns uniformly can cause noisy or irrelevant patterns to dominate the inference process. As shown in Table~\ref{tab:pitwi_ablation}, removing reweighting substantially increases token usage, which justifies the effectiveness of our online inductive learning method.  

Figure~\ref{fig:reveal_count_charts} further illustrates how these efficiency gains emerge during online learning. Overall, these trends support the need for online inductive learning: the pattern library becomes increasingly useful as more task trajectories are accumulated and put into learning.

\subsection{Qualitative Results and Out-of-Distribution Generalization}

Finally, we include a preliminary OOD generalizability study in Appendix~\ref{app:ood_generalization}. The result suggests that patterns and associated weights learned on smaller maps can zero-shot transfer to larger maps. Furthermore, we provide qualitative illustrations of pattern inference and induction in Fig.~\ref{fig:qualitative_craft}~\ref{fig:qualitative_frozenlake}\ref{fig:qualitative_cube}.

\section{Limitations and Future Work}
\label{sec:future}
As a proof-of-concept research, our work focuses on deterministic visual planning benchmarks rather than real-world or stochastic environments. We also assume that the set of possible variable types and the bounding box of each visual variable are known in advance. In this sense, our method addresses only one part of the broader TWI efficiency problem: how to reduce grounding cost when each reveal operation returns correct visual information and the symbolic interface is already specified. We plan to extend PI-TWI to real-world and stochastic environments by synergizing imputation and reranking. Specifically, \emph{imputation} will continue to reduce grounding costs for high-certainty structural regularities (e.g., repeated architectural patterns). For noisier, non-deterministic regularities, \emph{reranking} will serve as an interpretable heuristic to prioritize exploration and planning without committing to uncertain facts. This dual approach allows PI-TWI to exploit diverse environmental patterns while maintaining robustness to uncertainty.

\section*{Acknowledgement}
This work was supported by National Natural Science Foundation of China (U23A20311,62206245).
\bibliographystyle{named}
\bibliography{mllm}

\begin{thebibliography}{}

\bibitem[\protect\citeauthoryear{Dellin and Srinivasa}{2016}]{dellin2016unifying}
Christopher Dellin and Siddhartha Srinivasa.
\newblock A unifying formalism for shortest path problems with expensive edge evaluations via lazy best-first search over paths with edge selectors.
\newblock In {\em Proceedings of the international conference on automated planning and scheduling}, volume~26, pages 459--467, 2016.

\bibitem[\protect\citeauthoryear{Driess \bgroup \em et al.\egroup }{2023}]{driess2023palmeembodiedmultimodallanguage}
Danny Driess, Fei Xia, Mehdi S.~M. Sajjadi, Corey Lynch, Aakanksha Chowdhery, Brian Ichter, Ayzaan Wahid, Jonathan Tompson, Quan Vuong, Tianhe Yu, Wenlong Huang, Yevgen Chebotar, Pierre Sermanet, Daniel Duckworth, Sergey Levine, Vincent Vanhoucke, Karol Hausman, Marc Toussaint, Klaus Greff, Andy Zeng, Igor Mordatch, and Pete Florence.
\newblock Palm-e: An embodied multimodal language model, 2023.

\bibitem[\protect\citeauthoryear{Duan \bgroup \em et al.\egroup }{2025}]{duan2025got}
Chengqi Duan, Rongyao Fang, Yuqing Wang, Kun Wang, Linjiang Huang, Xingyu Zeng, Hongsheng Li, and Xihui Liu.
\newblock Got-r1: Unleashing reasoning capability of mllm for visual generation with reinforcement learning.
\newblock {\em arXiv preprint arXiv:2505.17022}, 2025.

\bibitem[\protect\citeauthoryear{Ellis \bgroup \em et al.\egroup }{2021}]{ellis2021dreamcoder}
Kevin Ellis, Catherine Wong, Maxwell Nye, Mathias Sabl{\'e}-Meyer, Lucas Morales, Luke Hewitt, Luc Cary, Armando Solar-Lezama, and Joshua~B Tenenbaum.
\newblock Dreamcoder: Bootstrapping inductive program synthesis with wake-sleep library learning.
\newblock In {\em Proceedings of the 42nd acm sigplan international conference on programming language design and implementation}, pages 835--850, 2021.

\bibitem[\protect\citeauthoryear{et al.}{2025}]{bai2025qwen3vltechnicalreport}
Shuai~Bai et~al.
\newblock Qwen3-vl technical report, 2025.

\bibitem[\protect\citeauthoryear{Evans and Grefenstette}{2018}]{evans2018learning}
Richard Evans and Edward Grefenstette.
\newblock Learning explanatory rules from noisy data.
\newblock {\em Journal of Artificial Intelligence Research}, 61:1--64, 2018.

\bibitem[\protect\citeauthoryear{Feng \bgroup \em et al.\egroup }{2025}]{feng2025reflectiveplanningvisionlanguagemodels}
Yunhai Feng, Jiaming Han, Zhuoran Yang, Xiangyu Yue, Sergey Levine, and Jianlan Luo.
\newblock Reflective planning: Vision-language models for multi-stage long-horizon robotic manipulation, 2025.

\bibitem[\protect\citeauthoryear{Fu \bgroup \em et al.\egroup }{2025}]{fu2025refocus}
Xingyu Fu, Minqian Liu, Zhengyuan Yang, John Corring, Yijuan Lu, Jianwei Yang, Dan Roth, Dinei Florencio, and Cha Zhang.
\newblock Refocus: Visual editing as a chain of thought for structured image understanding.
\newblock {\em arXiv preprint arXiv:2501.05452}, 2025.

\bibitem[\protect\citeauthoryear{Gao \bgroup \em et al.\egroup }{2025}]{gao2025cubebench}
Huan-ang Gao, Zikang Zhang, Tianwei Luo, Kaisen Yang, Xinzhe Juan, Jiahao Qiu, Tianxing Chen, Bingxiang He, Hao Zhao, Hao Zhou, et~al.
\newblock Cubebench: Diagnosing interactive, long-horizon spatial reasoning under partial observations.
\newblock {\em arXiv preprint arXiv:2512.23328}, 2025.

\bibitem[\protect\citeauthoryear{Garrett \bgroup \em et al.\egroup }{2021}]{garrett2021integrated}
Caelan~Reed Garrett, Rohan Chitnis, Rachel Holladay, Beomjoon Kim, Tom Silver, Leslie~Pack Kaelbling, and Tom{\'a}s Lozano-P{\'e}rez.
\newblock Integrated task and motion planning.
\newblock {\em Annual review of control, robotics, and autonomous systems}, 4(1):265--293, 2021.

\bibitem[\protect\citeauthoryear{{Google DeepMind}}{2026}]{google2026gemini31pro}
{Google DeepMind}.
\newblock Gemini 3.1 pro model card.
\newblock \url{https://deepmind.google/models/model-cards/gemini-3-1-pro/}, 2026.

\bibitem[\protect\citeauthoryear{Hafner}{2022}]{hafner2021benchmarking}
Danijar Hafner.
\newblock Benchmarking the spectrum of agent capabilities.
\newblock In {\em The Tenth International Conference on Learning Representations}, 2022.

\bibitem[\protect\citeauthoryear{He \bgroup \em et al.\egroup }{2022}]{he2022masked}
Kaiming He, Xinlei Chen, Saining Xie, Yanghao Li, Piotr Doll{\'a}r, and Ross Girshick.
\newblock Masked autoencoders are scalable vision learners.
\newblock In {\em Proceedings of the IEEE/CVF conference on computer vision and pattern recognition}, pages 16000--16009, 2022.

\bibitem[\protect\citeauthoryear{Hsu \bgroup \em et al.\egroup }{2025}]{hsu2025programs}
Joy Hsu, Emily Jin, Jiajun Wu, and Niloy~J Mitra.
\newblock From programs to poses: Factored real-world scene generation via learned program libraries.
\newblock {\em arXiv preprint arXiv:2510.10292}, 2025.

\bibitem[\protect\citeauthoryear{Hu \bgroup \em et al.\egroup }{2024}]{hu2024visual}
Yushi Hu, Weijia Shi, Xingyu Fu, Dan Roth, Mari Ostendorf, Luke Zettlemoyer, Noah~A. Smith, and Ranjay Krishna.
\newblock Visual sketchpad: Sketching as a visual chain of thought for multimodal language models.
\newblock In {\em The Thirty-eighth Annual Conference on Neural Information Processing Systems}, 2024.

\bibitem[\protect\citeauthoryear{Jacobs \bgroup \em et al.\egroup }{1991}]{jacobs1991adaptive}
Robert~A Jacobs, Michael~I Jordan, Steven~J Nowlan, and Geoffrey~E Hinton.
\newblock Adaptive mixtures of local experts.
\newblock {\em Neural computation}, 3(1):79--87, 1991.

\bibitem[\protect\citeauthoryear{Jones \bgroup \em et al.\egroup }{2023}]{jones2023shapecoder}
R~Kenny Jones, Paul Guerrero, Niloy~J Mitra, and Daniel Ritchie.
\newblock Shapecoder: Discovering abstractions for visual programs from unstructured primitives.
\newblock {\em ACM Transactions on Graphics (TOG)}, 42(4):1--17, 2023.

\bibitem[\protect\citeauthoryear{Jones \bgroup \em et al.\egroup }{2025}]{jones2025shapelib}
R~Kenny Jones, Paul Guerrero, Niloy~J Mitra, and Daniel Ritchie.
\newblock Shapelib: Designing a library of programmatic 3d shape abstractions with large language models.
\newblock {\em arXiv preprint arXiv:2502.08884}, 2025.

\bibitem[\protect\citeauthoryear{Kociemba}{2020}]{kociemba}
Herbert Kociemba.
\newblock Kociemba's solver.
\newblock \url{https://kociemba.org/}, 2020.

\bibitem[\protect\citeauthoryear{Li \bgroup \em et al.\egroup }{2025a}]{li2025imagine}
Chengzu Li, Wenshan Wu, Huanyu Zhang, Yan Xia, Shaoguang Mao, Li~Dong, Ivan Vuli{\'c}, and Furu Wei.
\newblock Imagine while reasoning in space: Multimodal visualization-of-thought.
\newblock {\em arXiv preprint arXiv:2501.07542}, 2025.

\bibitem[\protect\citeauthoryear{Li \bgroup \em et al.\egroup }{2025b}]{li2025dyfo}
Geng Li, Jinglin Xu, Yunzhen Zhao, and Yuxin Peng.
\newblock Dyfo: A training-free dynamic focus visual search for enhancing lmms in fine-grained visual understanding.
\newblock In {\em Proceedings of the Computer Vision and Pattern Recognition Conference}, pages 9098--9108, 2025.

\bibitem[\protect\citeauthoryear{Liu \bgroup \em et al.\egroup }{2024}]{liu2024enhancing}
Jingming Liu, Yumeng Li, Boyuan Xiao, Yichang Jian, Ziang Qin, Tianjia Shao, Yao-Xiang Ding, and Kun Zhou.
\newblock Enhancing visual reasoning with autonomous imagination in multimodal large language models.
\newblock {\em arXiv preprint arXiv:2411.18142}, 2024.

\bibitem[\protect\citeauthoryear{Luo \bgroup \em et al.\egroup }{2025}]{luo2025visual}
Tiange Luo, Lajanugen Logeswaran, Justin Johnson, and Honglak Lee.
\newblock Visual test-time scaling for gui agent grounding.
\newblock {\em arXiv preprint arXiv:2505.00684}, 2025.

\bibitem[\protect\citeauthoryear{Mei \bgroup \em et al.\egroup }{2024}]{mei2024replanvlmreplanningrobotictasks}
Aoran Mei, Guo-Niu Zhu, Huaxiang Zhang, and Zhongxue Gan.
\newblock Replanvlm: Replanning robotic tasks with visual language models, 2024.

\bibitem[\protect\citeauthoryear{{OpenAI}}{2025}]{openai2025thinkingwithimages}
{OpenAI}.
\newblock Thinking with images.
\newblock \url{https://openai.com/index/thinking-with-images/}, 2025.

\bibitem[\protect\citeauthoryear{{OpenAI}}{2026}]{openai2026gpt54}
{OpenAI}.
\newblock Introducing gpt-5.4.
\newblock \url{https://openai.com/index/introducing-gpt-5-4/}, 2026.

\bibitem[\protect\citeauthoryear{Piriyakulkij \bgroup \em et al.\egroup }{2025}]{piriyakulkij2025poe}
Wasu~Top Piriyakulkij, Yichao Liang, Hao Tang, Adrian Weller, Marta Kryven, and Kevin Ellis.
\newblock Poe-world: Compositional world modeling with products of programmatic experts.
\newblock {\em arXiv preprint arXiv:2505.10819}, 2025.

\bibitem[\protect\citeauthoryear{Shindo \bgroup \em et al.\egroup }{2023}]{shindo2023alpha}
Hikaru Shindo, Viktor Pfanschilling, Devendra~Singh Dhami, and Kristian Kersting.
\newblock $\alpha$ ilp: thinking visual scenes as differentiable logic programs.
\newblock {\em Machine Learning}, 112(5):1465--1497, 2023.

\bibitem[\protect\citeauthoryear{Shindo \bgroup \em et al.\egroup }{2024}]{shindo2024learning}
Hikaru Shindo, Viktor Pfanschilling, Devendra~Singh Dhami, and Kristian Kersting.
\newblock Learning differentiable logic programs for abstract visual reasoning.
\newblock {\em Machine Learning}, 113(11):8533--8584, 2024.

\bibitem[\protect\citeauthoryear{Su \bgroup \em et al.\egroup }{2025a}]{su2025openthinkimg}
Zhaochen Su, Linjie Li, Mingyang Song, Yunzhuo Hao, Zhengyuan Yang, Jun Zhang, Guanjie Chen, Jiawei Gu, Juntao Li, Xiaoye Qu, et~al.
\newblock Openthinkimg: Learning to think with images via visual tool reinforcement learning.
\newblock {\em arXiv preprint arXiv:2505.08617}, 2025.

\bibitem[\protect\citeauthoryear{Su \bgroup \em et al.\egroup }{2025b}]{su2025thinking}
Zhaochen Su, Peng Xia, Hangyu Guo, Zhenhua Liu, Yan Ma, Xiaoye Qu, Jiaqi Liu, Yanshu Li, Kaide Zeng, Zhengyuan Yang, et~al.
\newblock Thinking with images for multimodal reasoning: Foundations, methods, and future frontiers.
\newblock {\em arXiv preprint arXiv:2506.23918}, 2025.

\bibitem[\protect\citeauthoryear{Surís \bgroup \em et al.\egroup }{2023}]{suris2023vipergpt}
Dídac Surís, Sachit Menon, and Carl Vondrick.
\newblock Vipergpt: Visual inference via python execution for reasoning.
\newblock In {\em ICCV}, pages 11854--11864, 2023.

\bibitem[\protect\citeauthoryear{Towers \bgroup \em et al.\egroup }{2024}]{towers2024gymnasium}
Mark Towers, Ariel Kwiatkowski, Jordan Terry, John~U Balis, Gianluca De~Cola, Tristan Deleu, Manuel Goul{\~a}o, Andreas Kallinteris, Markus Krimmel, Arjun KG, et~al.
\newblock Gymnasium: A standard interface for reinforcement learning environments.
\newblock {\em arXiv preprint arXiv:2407.17032}, 2024.

\bibitem[\protect\citeauthoryear{Wang \bgroup \em et al.\egroup }{2025}]{wang2025visuothink}
Yikun Wang, Siyin Wang, Qinyuan Cheng, Zhaoye Fei, Liang Ding, Qipeng Guo, Dacheng Tao, and Xipeng Qiu.
\newblock Visuothink: Empowering lvlm reasoning with multimodal tree search.
\newblock In {\em Proceedings of the 63rd Annual Meeting of the Association for Computational Linguistics (Volume 1: Long Papers)}, pages 21707--21719, 2025.

\bibitem[\protect\citeauthoryear{Wei \bgroup \em et al.\egroup }{2022}]{wei2022chain}
Jason Wei, Xuezhi Wang, Dale Schuurmans, Maarten Bosma, Fei Xia, Ed~Chi, Quoc~V Le, Denny Zhou, et~al.
\newblock Chain-of-thought prompting elicits reasoning in large language models.
\newblock {\em Advances in neural information processing systems}, 35:24824--24837, 2022.

\bibitem[\protect\citeauthoryear{Wu and Xie}{2024}]{wu2024v}
Penghao Wu and Saining Xie.
\newblock V?: Guided visual search as a core mechanism in multimodal llms.
\newblock In {\em Proceedings of the IEEE/CVF Conference on Computer Vision and Pattern Recognition}, pages 13084--13094, 2024.

\bibitem[\protect\citeauthoryear{Wu \bgroup \em et al.\egroup }{2025}]{wu2025dimo}
Hang Wu, Hongkai Chen, Yujun Cai, Chang Liu, Qingwen Ye, Ming-Hsuan Yang, and Yiwei Wang.
\newblock Dimo-gui: Advancing test-time scaling in gui grounding via modality-aware visual reasoning.
\newblock {\em arXiv preprint arXiv:2507.00008}, 2025.

\bibitem[\protect\citeauthoryear{Xu \bgroup \em et al.\egroup }{2026}]{xu2026visual}
Yi~Xu, Chengzu Li, Han Zhou, Xingchen Wan, Caiqi Zhang, Anna Korhonen, and Ivan Vuli{\'c}.
\newblock Visual planning: Let's think only with images.
\newblock In {\em The Fourteenth International Conference on Learning Representations}, 2026.

\bibitem[\protect\citeauthoryear{Zhao \bgroup \em et al.\egroup }{2025}]{zhao2025cotvlavisualchainofthoughtreasoning}
Qingqing Zhao, Yao Lu, Moo~Jin Kim, Zipeng Fu, Zhuoyang Zhang, Yecheng Wu, Zhaoshuo Li, Qianli Ma, Song Han, Chelsea Finn, Ankur Handa, Ming-Yu Liu, Donglai Xiang, Gordon Wetzstein, and Tsung-Yi Lin.
\newblock Cot-vla: Visual chain-of-thought reasoning for vision-language-action models, 2025.

\bibitem[\protect\citeauthoryear{Zheng \bgroup \em et al.\egroup }{2026}]{zheng2026deepeyes}
Ziwei Zheng, Michael Yang, Jack Hong, Chenxiao Zhao, Guohai Xu, Le~Yang, Chao Shen, and XingYu.
\newblock Deepeyes: Incentivizing ''thinking with images'' via reinforcement learning.
\newblock In {\em The Fourteenth International Conference on Learning Representations}, 2026.

\end{thebibliography}

\appendix

\section{More Experimental Results}

\subsection{OOD Generalization}
\label{app:ood_generalization}

In short: patterns naturally share across different sizes of the same kind of experiments, but weights don't always.

We want to test if, after getting trained on smaller maps, the patterns and weights could zero-shot generalize to larger maps. This means that not only are the patterns fixed, but also the weights are fixed. And they will not be trained on larger maps.

We tested \textsc{Crafter}'s learned pattern \& weights from 64x64 map on a larger 128x128 map. The result shows that patterns \& their weights have strong OOD generalization capability. The reveal count drops from 2349.63 to 1570.45, indicating its effectiveness.

\subsection{More Qualitative Results}
\begin{figure}[h]
    \centering
    \includegraphics[width=.8\textwidth]{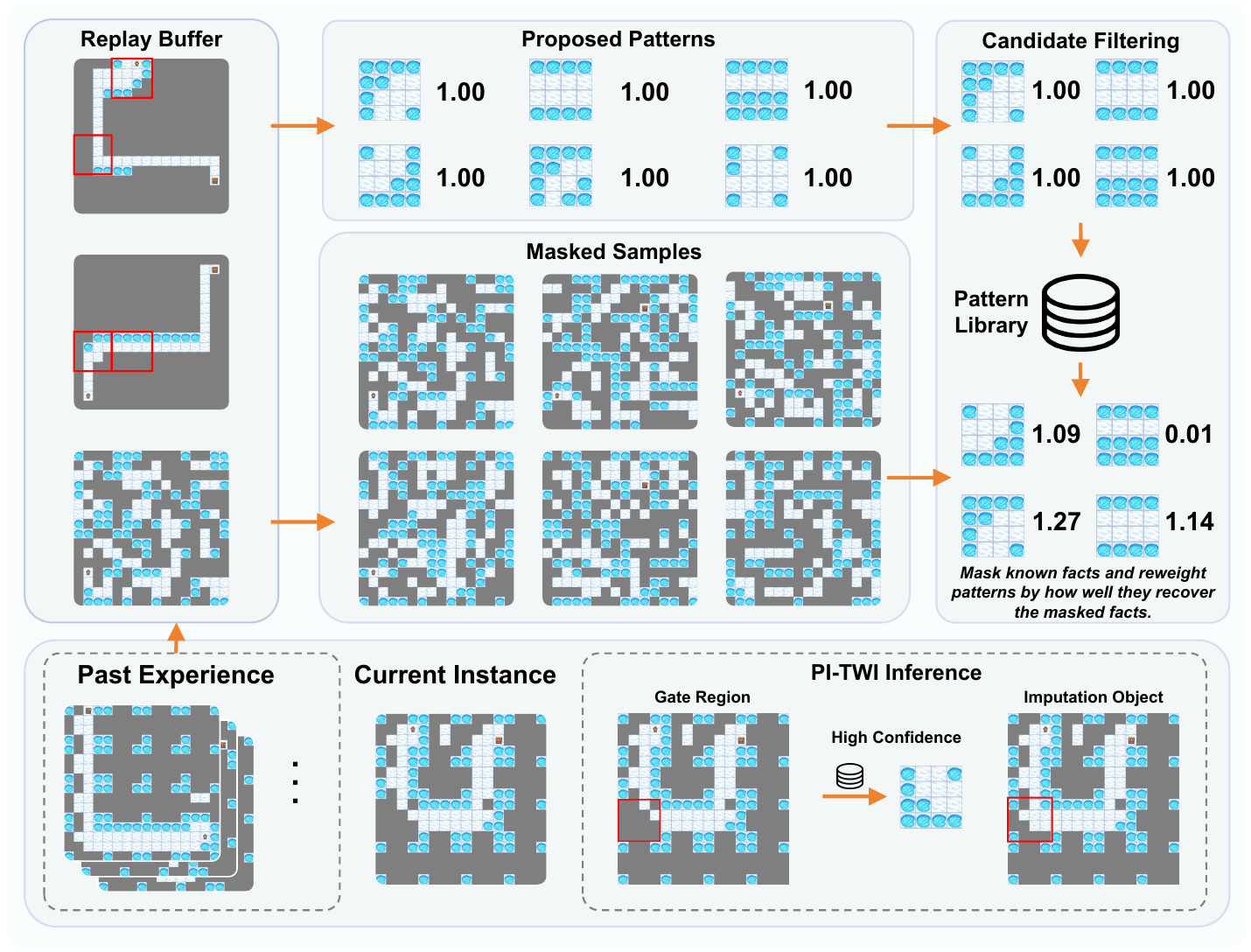}
    \caption{Qualitative illustration of the pattern inference and induction processes in \textsc{FrozenLake}.} 
    \label{fig:qualitative_frozenlake} 
\end{figure}

\begin{figure}[h]
    \centering
    \includegraphics[width=.8\textwidth]{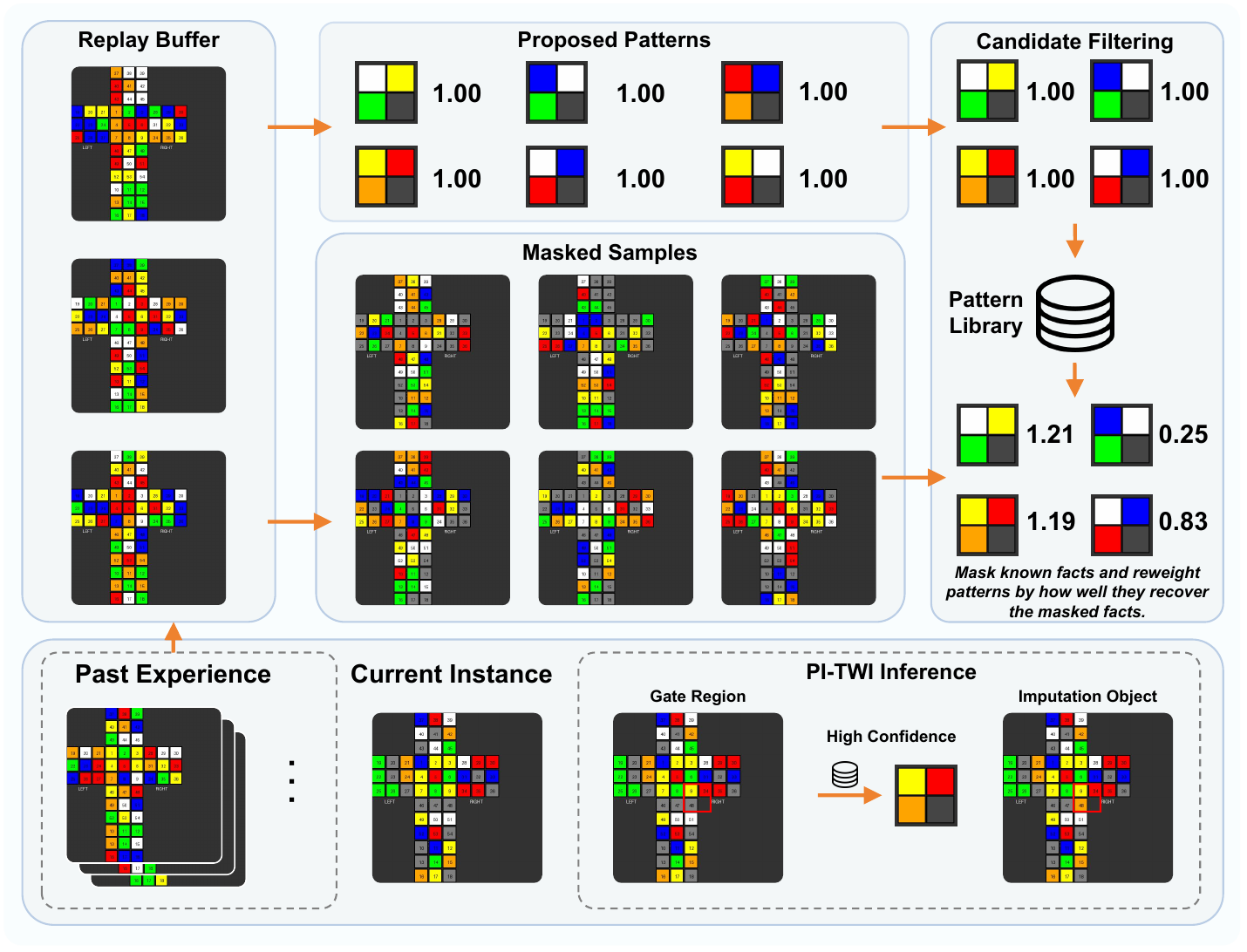}
    \caption{Qualitative illustration of the pattern inference and induction processes in \textsc{CubeBench}.} 
    \label{fig:qualitative_cube} 
\end{figure}

Qualitative illustration of the online inductive learning process under \textsc{FrozenLake} and \textsc{CubeBench} are illustrated in Fig.~\ref{fig:qualitative_frozenlake} and Fig.~\ref{fig:qualitative_cube}.
\section{Implementation Details}
\label{app:implementation}

Our experiments are conducted on a personal computer with Mac M3 Max CPU (16 cores) and 128GB RAM. For VLM calls, we utilize OpenRouter\footnote{https://openrouter.ai/} and OpenAI API Platform\footnote{https://platform.openai.com/}.

We also modify \textsc{Crafter} and \textsc{Frozenlake} environments, which are both built upon Gymnasium~\citep{towers2024gymnasium} framework, to fit our purposes.

There are 100 episodes for each run on each benchmark. For each benchmark, we conduct a total of 9 runs. Specifically, we employ 3 seeds to initialize distinct environment instances (map or state sequences). For each seed, 3 trials are performed to ensure statistical robustness. The results in Tab.~\ref{tab:pitwi_baseline} and Tab.~\ref{tab:pitwi_ablation} are mean results over runs.

\paragraph{Relation between token consumption and reveal count.}
\label{sec:token_reveal_count_relation}

We use total token consumption as the primary efficiency metric throughout the paper. Reveal count is reported only as a diagnostic quantity to explain where the reduction in \emph{perception} token consumption comes from. In our implementation, each reveal operation within the same task uses a fixed prompt template, image resolution, and output schema. We therefore account for perception-token usage with a fixed per-reveal token budget:

\[
T_{\mathrm{perc}} = c_{\mathrm{perc}} \cdot N_{\mathrm{perc}},
\]

where $T_{\text{perc}}$ is the perception token count, $N_{\mathrm{perc}}$ is the reveal count and $c_{\mathrm{perc}}$ is the task-specific token budget per reveal operation. The total token cost additionally includes proposal tokens, so all main efficiency comparisons are still reported using total token consumption.

For \textsc{FrozenLake} and \textsc{CubeBench}, the per-reveal perception costs are fixed: 98 input and 5 output tokens for \textsc{FrozenLake}, and 88 input and 5 output tokens for \textsc{CubeBench}. For \textsc{Crafter}, the input cost is fixed at 184 tokens, while the output cost can vary slightly from 3 to 5 tokens depending on the predicted cell type. We therefore use 5 output tokens as a conservative per-reveal budget for \textsc{Crafter}. This convention provides an upper-bound estimate of perception-token usage; since the variation is at most two output tokens per reveal and is negligible compared to the input-token cost, it does not affect the relative efficiency comparison.

An example: In PI-TWI's ``w/o inference'' ablation for \textsc{CubeBench}, the 4.75k input perception tokens come from 54 reveal operations (since a Rubik's cube has 54 facelets), namely $54 \times 88 = 4752$.

\subsection{Policy-generation procedure}

\textsc{FrozenLake} \quad This environment follows \textsc{LazySP}~\citep{dellin2016unifying} shortest path algorithm. Simply put, the \textsc{LazySP} algorithm is like this: at first, we only know the locations of the start and the goal, but not any passable ``ground'' or impassable ``hole''. We plan optimistically: 

\begin{enumerate}
    \item We assume every unrevealed grid is passable. On top of this, we plan the shortest path.
    \item If there aren't any unrevealed grids on the path, the algorithm stops and returns the path
    \item If there are, we check the unrevealed grids one by one from start to goal.
    \item If an impassable grid is ever encountered during this process, we immediately go back to step 1 to get another plan.
    \item If all checked grids are passable, the algorithm stops and returns the path.
\end{enumerate}

The \textit{policy-generation procedure} works at step (4) above. Each time, it outputs the first unrevealed grid from start to goal.

\begin{figure}[ht] 
    \centering
    \includegraphics[width=0.5\textwidth]{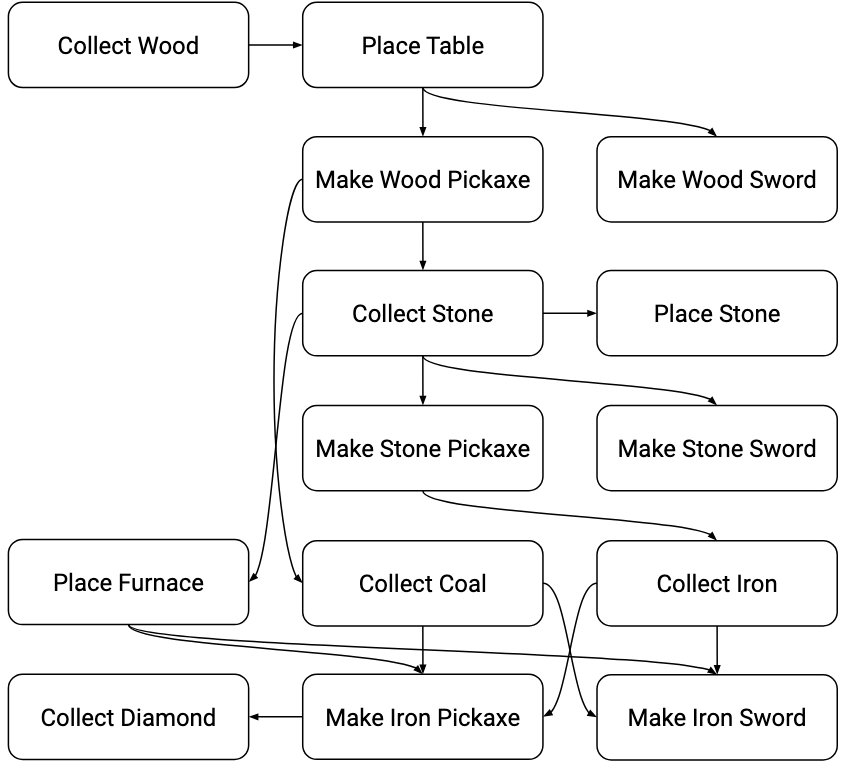}

    \caption{Achievements for \textsc{Crafter}} 
    \label{fig:achivements} 
\end{figure}

\textsc{Crafter} \quad This environment, just like the original Minecraft, contains many achievements, The success condition is to finish all 14 achievements in Fig.~\ref{fig:achivements}.

Inspired by task and motion planning, we treat each ``achievement'' as a \emph{task}. Since there are dependencies between these tasks (Fig.~\ref{fig:achivements}), we can get its topological order (see ~\ref{fig:crafter_topological_order}) and finish tasks one by one with this order.

And for each specific task, we have to gather enough materials to craft a certain tool. In order to gather materials, we have to first find the required materials for this specific task, and plan a feasible route to it. Here, we design an algorithm that do both at the same time. This is where the \emph{policy-generation procedure} $G$ works: it finds all passable grids among the revealed grids~\footnote{``grass'', ``sand'' and ``path'' are natural passable grids; ``tree'' is also passable, since you can mine it with your bare hands and pass it; if you have wooden pickaxe, ``stone'' and ``coal'' will also become passable, since they are now minable; etc. Basically, a grid being passable means you can either directly walk through it, or do so after using your existing tools like pickaxe.}, and outputs all their \emph{adjacent unrevealed grid}. Intuitively, this $G$ gradually expand the boundary of revealed grids, while making sure all passable revealed grids are connected. In this way, we can make sure if we find a certain type of material, we can have at least one feasible route towards it.  

\textsc{CubeBench} We adopt the environment and solver interface from CubeBench~\citep{gao2025cubebench}. In this setting, the input is an image observation of a scrambled Rubik's Cube, and the VLM is asked to convert the visual observation into a complete 54-character symbolic state. Each character denotes one sticker color using the CubeBench color alphabet R, G, B, Y, O, and W, ordered according to the CubeBench simulator convention. 

After the VLM outputs the 54-character state code, the code is passed to the CubeBench solver interface. The solver is used as a black-box symbolic backend that returns a sequence of cube moves for the predicted state. Our focus is therefore on obtaining an accurate symbolic state from the image before invoking the solver, not on modifying or analyzing the solver's internal search procedure.

\subsection{Pattern Proposal}

Since \textsc{FrozenLake} is 16x16 and \textsc{Crafter} is 64x64, if you put the entire symbolic world model into the prompt, it will contain too many tokens, and VLMs will fail to generate good proposals in such long context.

So, we need to preprocess the symbolic world models from the replay buffer.

\textsc{FrozenLake}
\begin{enumerate}
    \item We partition each 16x16 map in the replay buffer into 16 non-overlapping 4x4 minimaps.
    \item Minimaps containing fewer than 8 revealed grids are discarded to ensure sufficient information density.
    \item We randomly sample 5 minimaps from the remaining candidates.
    \item These 5 minimaps are subsequently integrated into our prompt template.
\end{enumerate}

This proposal will be triggered every 5 maps. So for in total 100 episodes, there will be 19 proposals.

\textsc{Crafter}
\begin{enumerate}
    \item For each 64x64 map in the replay buffer, we identify the minimal bounding box enclosing all revealed grids and subsequently crop the map to these boundaries.
    \item Each cropped map is subdivided into multiple 15x15 minimaps. To accommodate cross-shaped patterns, these minimaps maintain at least a 3x3 overlap. We optimize the minimap locations such that the maximum overlap is minimized.
    \item We randomly sample 5 minimaps from the resulting set.
    \item These 5 minimaps are integrated into our prompt template.
\end{enumerate}

This proposal will be triggered every 5 maps. So for in total 100 episodes, there will be 19 proposals.

\textsc{CubeBench} 
\begin{enumerate}
    \item We select unreflected states from the replay buffer, prioritizing failed reconstructions and using successful reconstructions only when additional examples are needed.
    \item For each selected state, we extract the eight corner cubies and represent each corner by its three-color token.
    \item We add the corner cubie-to-sticker-index mapping and a deduplicated summary of observed corner tokens with their multiplicities.
    \item These examples are integrated into our prompt template, and the model proposes single-corner patterns in strict JSON format.
\end{enumerate}

This proposal will be triggered every 10 episodes. So for in total 100 episodes, there will be 9 proposals, since no proposal is needed after the final episode.

The prompts are listed below. The minimaps will be filled into each ``\{PARTIAL\_MAP\_BLOCK\}'' in the prompt. The ``\{K\}'' will be 10 for \textsc{Crafter}, 10 for \textsc{FrozenLake} and 3 for \textsc{CubeBench}.

\begin{promptbox}{Prompt for FrozenLake Skill Proposal}
\begin{lstlisting}[style=promptstyle]
### Task Description
You will be provided with symbolic partial maps or aligned tiles from a FrozenLake-style environment.

Each map uses three labels:
- SAFE: observed traversable cell
- HOLE: observed unsafe cell
- UNKNOWN: unobserved cell

Your goal is to infer up to {K} reusable 4x4 local navigation skill patterns from the provided examples.

### Grid Alignment Rules
Full maps are divided into non-overlapping aligned 4x4 blocks; smaller provided examples may already be aligned tiles.

Only these aligned 4x4 blocks may be considered.
Do not use sliding windows.
Do not shift, crop, rotate, resize, or merge blocks.

### Pattern Guidelines
* Look for paths: Ideal local patterns resemble safe structures like corridors or turns surrounded by holes.
* Prioritize reusability: Look for common, generalizable shapes rather than copying one-off observations.
* No unknowns: Every cell in your final patterns must be labeled exactly "SAFE" or "HOLE".

Note even if there are "UNKNOWN"s in the examples, you are still pretty much able to infer and guess.

### Strict Output Format
Return your answer as valid JSON ONLY. Do not provide conversational text or Markdown code blocks outside the JSON.

Structure:
{
  "patterns": [
    [
      ["R1C1", "R1C2", "R1C3", "R1C4"],
      ["R2C1", "R2C2", "R2C3", "R2C4"],
      ["R3C1", "R3C2", "R3C3", "R3C4"],
      ["R4C1", "R4C2", "R4C3", "R4C4"]
    ]
  ]
}

### Example {i}
{PARTIAL_MAP_BLOCK}
\end{lstlisting}
\end{promptbox}
\begin{promptbox}{Prompt for \textsc{Crafter} Pattern Proposal}
\begin{lstlisting}[style=promptstyle]
You are given partially revealed Crafter material maps.
Crafter is essentially 2D Minecraft; so also use that world knowledge to infer.
Each pattern should use top/bottom/left/right to predict the center.
Propose at most {K} reusable cross-shaped patterns.
Return strict JSON only with shape {"patterns": [{"center": "...", "top": "...", "bottom": "...", "left": "...", "right": "..."}]}.
Use only lowercase world-generation material names. Do not use unknown.

Example {i}
episode={episode_id}
bbox: x=[{bbox_xmin},{bbox_xmax}), y=[{bbox_ymin},{bbox_ymax})
{PARTIAL_MAP_BLOCK}
\end{lstlisting}
\end{promptbox}
\begin{promptbox}{Prompt for \textsc{CubeBench} Pattern Proposal}
\begin{lstlisting}[style=promptstyle]
You are extracting reusable corner-cubie skills from pre-extracted corner observations.
Each replay example already contains all 8 corner cubies (CORNERS_8).
Character alphabet for each corner token is exactly {R,G,B,Y,O,W}.
Only use corner cubies: {corner_cubie_names}.
Corner cubie-to-sticker mapping:
{corner_cubie_to_sticker_mapping}

Task:
- Read CORNERS_8 examples.
- Extract fully-specified corner tokens as <CUBIE>:<TOKEN>.
- Consolidate duplicates by identical <CUBIE>:<TOKEN> pairs (same pair can appear in multiple examples).
- Each proposed pattern must contain exactly one corner cubie.
- Corner token length must be exactly 3.

Replay examples:
{replay_examples}

Deduplicated corner token summary (with multiplicity):
{deduplicated_corner_token_summary}

Propose at most {patterns_per_trigger} new reusable skills.
Return strict JSON only with shape:
{"patterns": [{"cubies": {"URF": "ROW"}}, ...]}
Do not add commentary. Do not use markdown.
```

Here `{corner_cubie_names}` is the list of corner cubie identifiers, `{corner_cubie_to_sticker_mapping}` lists each corner cubie's simulator-order sticker indices with face names, `{replay_examples}` contains the selected `CORNERS_8` states, and `{deduplicated_corner_token_summary}` contains the unique `<CUBIE>:<TOKEN>` pairs with multiplicities.
\end{lstlisting}
\end{promptbox}

\subsection{Imputation}
\label{app:imputation}

In addition to the pattern-based imputation rule introduced in Sec.~\ref{sec:method_patterns}, there are two hyperparameters worth noting: the confidence threshold $\tau$ and the smoothing constant $\varepsilon$. The threshold $\tau$ controls how conservative the agent is when trying to add pattern-imputed facts into $K_t^I$. A larger $\tau$ reduces the risk of imputing incorrect symbolic facts into the world model. The smoothing constant $\varepsilon$, which is originally used in Eq.~\ref{eq:pattern_expert} to avoid assigning zero probability to non-predicted values, can also affect the conservativeness. We use a fixed $\varepsilon=0.001$ across all domains.

\begin{center}
\begin{tabular}{lcc}
\toprule
Domain & $\tau$ & $\varepsilon$ \\
\midrule
\textsc{FrozenLake} & $0.99$ & $0.001$ \\
\textsc{Crafter} & $1.00$ & $0.001$ \\
\textsc{CubeBench} & $0.99$ & $0.001$ \\
\bottomrule
\end{tabular}
\end{center}

\subsection{Reranking}
\label{app:reranking}

Reranking uses the same pattern-predicted distribution $p_{\boldsymbol w}(y_u\mid M_t)$ as imputation, but it does not add any unrevealed facts to the world model. Instead, it only changes the order in which candidate variables from $G(M_t)$ are revealed. Therefore, reranking can exploit stochastic regularities without just imputing them as hard symbolic facts.

In our experiments, reranking is used in \textsc{Crafter}, where $G(M_t)$ often returns many adjacent unrevealed visual variables on the current frontier. For a resource-acquisition subtask with the set of target resource types $Z$, we score each candidate variable $u \in G(M_t)$ by
\[
    s(u) = p_{\boldsymbol w}(y_u\in Z \mid M_t),
\]
and reveal the candidate with the highest score. This allows the agent to spend perception calls preferentially on variables that are more likely to contain task-relevant resources, such as trees, stone, coal, iron, or diamond, while still requiring direct perception before any such variable is added to the world model.

\subsection{Weight Optimization}

Since the max number of patterns in the library is relatively small, we choose L-BFGS as our gradient-based optimization method for faster convergence.

For \textsc{Frozenlake}, the learning rate is $1.0$ and the maximum iteration is 50.
For \textsc{Crafter}, the learning rate is $1.0$ and the maximum iteration is 20.
For \textsc{CubeBench}, the learning rate is $1.0$ and the maximum iteration is 150.

\section{Benchmark Details}
\label{app:bench}
We introduce three benchmarks in our experiments. (1) {\bf FrozenLake}~\citep{towers2024gymnasium}. The agent observes a rendered grid world and must find a shortest safe path from the start to the goal without entering holes. Each visual variable is a grid cell, and its value is the cell type. (2) {\bf Crafter.}~\citep{hafner2021benchmarking}. \textsc{Crafter} is originally a stochastic survival environment. We repurpose it as a deterministic visual resource acquisition and crafting task: the PI-TWI agent is given the task specification, and must ground enough of the map to plan a feasible sequence of navigation, collection, and crafting actions. Each visual variable is a grid cell, and its value is the corresponding terrain, object, or resource type. (3) {\bf CubeBench.}~\citep{gao2025cubebench}. We evaluate cube-state grounding as a structured visual planning problem. Each instance renders a cube state; the agent must reconstruct enough symbolic facelet information for a downstream cube planner or solver. Each visual variable is a facelet and its value is one of the six colors. 

There are 100 episodes for each run on each benchmark.

We repurpose \textsc{FrozenLake} and \textsc{Crafter} to make them suitable for our method.

\subsection{\textsc{FrozenLake}}

Action space: move up, move down, move left, move right.

Originally, for each grid in this environment, it has a fixed probability to become a ``hole''. So the map is randomly generated and does not have any patterns.

\begin{figure}[t]
    \centering
    \begin{tabular}{cccccc}
        \includegraphics[width=0.11\textwidth]{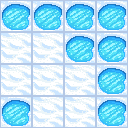} &
        \includegraphics[width=0.11\textwidth]{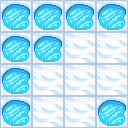} &
        \includegraphics[width=0.11\textwidth]{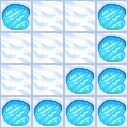} &
        \includegraphics[width=0.11\textwidth]{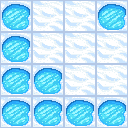} &
        \includegraphics[width=0.11\textwidth]{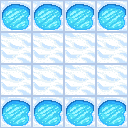} &
        \includegraphics[width=0.11\textwidth]{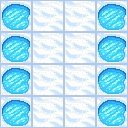} \\
        Pattern 1 & Pattern 2 & Pattern 3 & Pattern 4 & Pattern 5 & Pattern 6 \\
    \end{tabular}
    \caption{The six manually designed 4x4 macro patterns used to generate the modified \textsc{FrozenLake} maps.}
    \label{fig:frozenlake_patterns}
\end{figure}

In our modified version, we designed six 4x4 macro patterns (Fig. ~\ref{fig:frozenlake_patterns}). We also make sure the shortest path from start to goal is above a minimum. Then, we generate a map with this process:

\begin{enumerate}
    \item First divide the 16x16 map into 16 4x4 minimaps
    \item For each minimap, we randomly put one of six 4x4 macro patterns there
    \item We randomly sample the start and goal
    \item If the length of the shortest path in between is less than the minimum, we discard this map and go back to step (1); otherwise, we return this map.
\end{enumerate}

Our experiments are done with the minimal length of 25.

This generation method is basically rejection sampling, ensuring more uniform distribution. 

However, for our OOD test, where we need to generate 32x32 maps with minimal length of 50, the efficiency is too low, so we adopt another method:

\begin{enumerate}
    \item First divide the 16x16 map into 16 4x4 minimaps
    \item For each minimap, we randomly put one of six 4x4 macro patterns there
    \item We calculate the length of shortest path between all pairs of grids using \texttt{all\_pairs\_shortest\_path\_length} from \texttt{networkx}
    \item We filter all pairs whose lengths are smaller than minimum
    \item If there's no pairs left, we discard this map and go back to step (1); otherwise, we uniformly sample one pair as start and goal, and return the map.
\end{enumerate}

Essentially, this is still rejection sampling, but with much better efficiency.

\subsection{\textsc{Crafter}}

The original \textsc{Crafter} has 17 actions available. For our purpose, we removed action \texttt{Noop} and \texttt{Sleep}. The remaining actions are: move left, move right, move up, move down, do, place stone, place table, place furnace, place plant, make wood pickaxe, make stone pickaxe, make iron pickaxe, make wood sword, make stone sword and make iron sword.

Among the many game mechanics, we removed: (1) Spawned mobs (zombie, skeleton) and creatures (cow) (2) Moving objects (arrow), and disable: (1) Hunger (2) Thirst (3) Energy (4) Daylight cycle. We remove these two stochastic elements because stochasticity will make TWI for open-loop planning almost impossible. For similar reasons, we also disable exogenous mechanics like hunger and thirst.

There are originally 22 achievements. We remove 8 of them, which are:

\begin{enumerate}
    \item \textbf{Collect Drink}: Removed because the thirst mechanic is disabled.
    \item \textbf{Defeat Zombie \& Defeat Skeleton}: Removed because mob spawning is disabled.
    \item \textbf{Eat Cow}: Removed because both hunger mechanic and animal spawning are disabled.
    \item \textbf{Wake up}: Removed because the daylight cycle mechanic is disabled.
    \item \textbf{Collect Sapling, Place Plant, \& Eat Plant}: While these could technically be integrated into the planner, they do not serve our research goals. Since the player always spawns on grass, these tasks can be completed through repetitive actions in a single location (e.g., harvesting grass until a sapling drops). Because they require no environmental exploration or navigation, they offer no meaningful challenge to the agent's spatial perception.
\end{enumerate}

Since we no longer have to collect saplings, we also set the probability of getting a sapling from grass to zero.

The remaining 14 achievements will be done one by one in this order, which is one of the possible topological order of the DAG of the achievements:

\label{fig:crafter_topological_order}
\begin{center}
    \small
    collect wood $\rightarrow$ place table $\rightarrow$ make wood pickaxe $\rightarrow$ make wood sword $\rightarrow$ collect stone $\rightarrow$ place stone $\rightarrow$ make stone pickaxe $\rightarrow$ make stone sword $\rightarrow$ place furnace $\rightarrow$ collect coal $\rightarrow$ collect iron $\rightarrow$ make iron pickaxe $\rightarrow$ make iron sword $\rightarrow$ collect diamond
\end{center}

When generating a map, we will make sure each map gets enough resources to complete all 14 achievements. In total, we need 9 trees, 8 stones, 3 coal ores, 3 iron ores and 1 diamond ore. If one map is generated without enough resources, we will discard and regenerate the map.

\subsection{\textsc{CubeBench}}
The original CubeBench task takes an image observation of a scrambled Rubik's Cube, asks the model to recover the complete 54-character symbolic cube state, and then passes that state to the Kociemba's Solver that returns a sequence of cube moves. For our purpose, we keep the CubeBench state representation, color alphabet, simulator-order indexing, and solver interface, but remove the final cube-solving action sequence from the main evaluation target. Since an incorrect reconstructed state makes the downstream solve plan meaningless, we focus on the partial-observation state reconstruction problem before solver execution.

We replace one-shot visual state prediction with an active reveal protocol. A rollout begins with all 54 stickers unknown; each round actively reads one ground-truth sticker and then runs pattern-assisted imputation over the remaining unknown stickers. Inferred stickers do not count as active reveals. The experiment data is generated locally from CubeBench: the default configuration creates 100 states with 20 random scramble moves and seed 42, and saves the state list, scramble metadata, global cubie map, and sticker adjacency table. We fix the evaluated states and scramble seed so that performance differences can be attributed to active information gathering, pattern coverage, LLM-proposed skills, and learned weighting rather than dataset-generation stochasticity.

We evaluate two metrics: exact reconstruction accuracy, where all 54 predicted stickers must match the ground truth, and active reveal cost, which counts only stickers explicitly read from the ground truth. By default, the runner evaluates 100 states from the generated dataset. Pattern libraries are grouped by source, including ground-truth patterns, irrelevant patterns, placeholder patterns, and LLM-proposed patterns. The default placeholder library uses one-unknown corner templates with 24 unique candidates for each corner cubie position, and the current LLM proposal implementation accepts exactly one fully specified corner cubie per proposed pattern. The configuration also supports ablations that disable inference, reweighting, placeholders, ground-truth patterns, or LLM proposal, separating the effects of active imputation, placeholder coverage, proposal, and learned weighting.

\subsection{Macro Patterns}

A macro pattern in \textsc{FrozenLake} can be seen in Fig.~\ref{fig:frozenlake_patterns}. One macro patterns corresponds to 16 patterns, where the prediction of each pattern is one of the grid among the $4\times4=16$ grids, and the gate of each pattern is composed of other 15 grids.

Similarly a macro pattern of \text{CubeBench} corresponds to 3 patterns, where the prediction of each pattern is one of the facelet among the 3 facelets, and the gate of each pattern is composed of other 2 facelets.

The macro pattern of \textsc{Crafter} corresponds to only one pattern, where the prediction is the center grid and the gate is composed of the four surrounding it.

\subsection{Ground-Truth Planner}
\label{app:planner}
In each domain, the ground-truth planner is instantiated by LazySP~\citep{dellin2016unifying} for \textsc{FrozenLake}, by a task-and-motion-planning-style search procedure~\citep{garrett2021integrated} for \textsc{Crafter}, and by Kociemba's Solver~\citep{kociemba} for \textsc{CubeBench}. Note that in \textsc{FrozenLake}, the planner and grounding controller $G$ are the same due to the native grounding guidance ability of the planners.

\section{Baseline Details}
\label{app:baseline}
\subsection{Direct VLM Output}

For the ``Direct VLM Output'' experiment, prompts are provided directly to the model to elicit feedback. Note that while the first image in each prompt is an episode-specific map that varies, all other images in the prompts are static and consistent across experiments.

\subsection{Native TWI}

For the ``Native TWI'' experiment, prompts are given to the native TWI agent for each model. For GPT-5.4, the agent is called by ``code interpreter'' via OpenAI API; for Gemini 3.1 Pro, it is ``Code execution''; for Qwen3 VL 235B A22B, it is the official ``think\_with\_images'' Jupyter notebook of QwenLM/Qwen3-VL repository on Github. The prompts are similar to that of ``Direct VLM Output''. The main difference is that here we ban it from using any tools other than those for the purpose of cropping images, since other methods like using Python programs to analyze the pixel of a grid and output its type aren't in our consideration. If the agent does use Python in a specific task, it will be considered as plagiarism and fail this task.

\section{Potential Societal Impact}
\label{app:impact}
Our work is purely about improving visual planning abilities of VLMs. We do not find any significant societal impact requiring to be explicitly discussed.

\section{Prompts for Baselines}
We provide the prompts for the baseline methods of direct VLM output and native TWI.

\begin{promptbox}{Prompt for \textsc{FrozenLake} Direct VLM Output}
\begin{center}
    \includegraphics[width=0.5\textwidth]{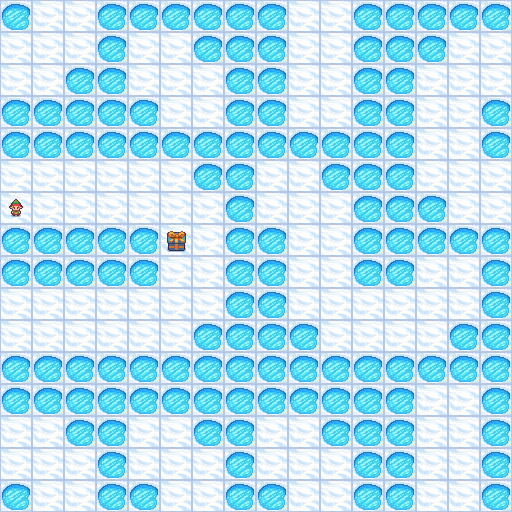}
    \includegraphics[width=0.5\textwidth]{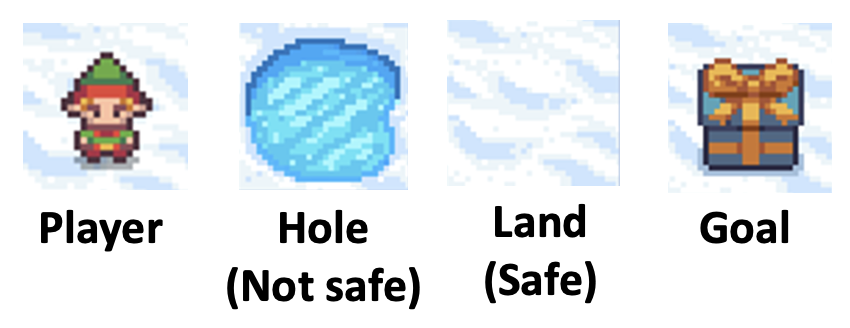}
\end{center}
\begin{lstlisting}[style=promptstyle]
Output the 16x16 map of this image, with the help of the reference. Suppose left-top is (0,0), and bottom-right is (15, 15). (row, col)  

You only need to output all coords of holes.

Format:
  
```json  
{"Holes": [(..., ...), ...]}
```
\end{lstlisting}
\end{promptbox}
\begin{promptbox}{Prompt for \textsc{Crafter} Direct VLM Output}
\begin{center}
    \includegraphics[width=0.5\textwidth]{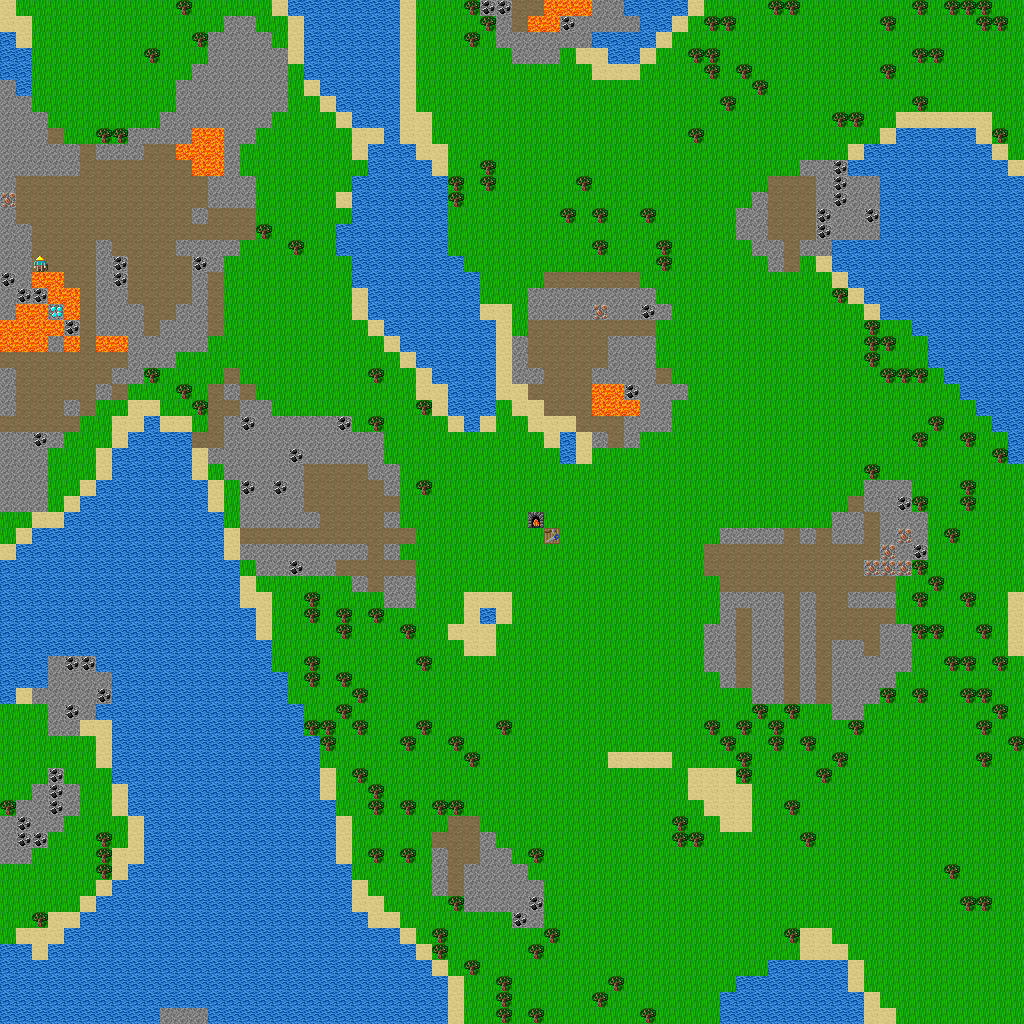}
    \includegraphics[width=0.5\textwidth]{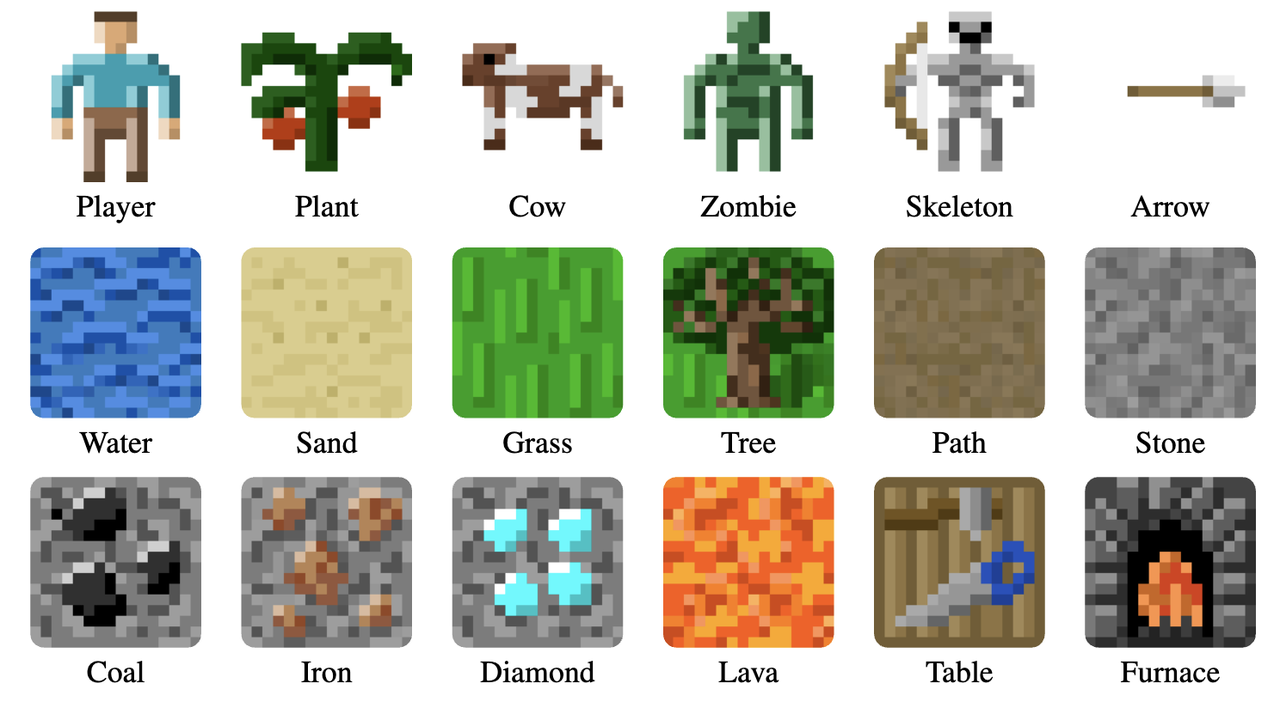}
\end{center}
\begin{lstlisting}[style=promptstyle]
Output the 64x64 map of this image, with the help of the reference. Suppose left-top is (0,0), and bottom-right is (63, 63). (row, col)  
  
Format:
  
```json  
{"sand": [(..., ...), ...], "grass": [...], ...}  
```
\end{lstlisting}
\end{promptbox}
\begin{promptbox}{Prompt for \textsc{CubeBench} Direct VLM Output}
\begin{center}
    \includegraphics[width=0.5\textwidth]{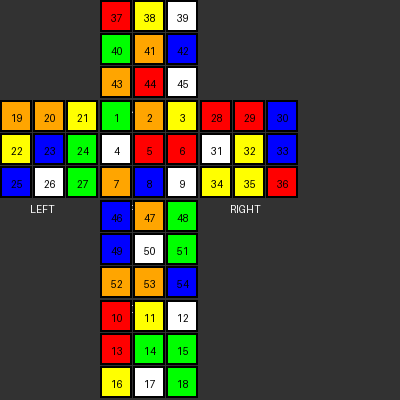}
\end{center}
\begin{lstlisting}[style=promptstyle]
You are a Rubik's Cube state reader. The image shows each sticker with a global numeric ID.

Read sticker colors by ID from 1 to 54 and build `state` as this exact ID order.

[Color codes]
Only use: R, G, B, Y, O, W

[Strict output requirements]
* Output only one JSON object
* No explanation, no markdown, no extra text
* state must be exactly 54 characters
* state must be exactly colors of ID 1..54 in order
* state characters must be only R/G/B/Y/O/W

[Allowed output format]
{"state":"XXXXXXXXXXXXXXXXXXXXXXXXXXXXXXXXXXXXXXXXXXXXXXXXXXXXXX"}
\end{lstlisting}
\end{promptbox}
\begin{promptbox}{Prompt for \textsc{FrozenLake} Native TWI}
\begin{center}
    \includegraphics[width=0.5\textwidth]{prompts/direct_vlm_output/frozenlake-1.png}
    \includegraphics[width=0.5\textwidth]{prompts/direct_vlm_output/frozenlake-2.png}
\end{center}
\begin{lstlisting}[style=promptstyle]
Output the 16x16 map of this image, with the help of the reference. Suppose left-top is (0,0), and bottom-right is (15, 15). (row, col)  

You only need to output all coords of holes.

Format:
  
```json  
{"Holes": [(..., ...), ...]}
```
  
Warning: You cannot use any tools other than vanilla "image cropping" to analyze the image. You are especially banned from using any coding besides the simplest cropping without any further visual cues.

This means, besides basic system commands like `cd`, `mkdir`, `ls`, etc., the only Bash command you can use is `magick` to crop the image. Also, you are completely banned from running `python` or any kind of popular programming language script.
\end{lstlisting}
\end{promptbox}
\begin{promptbox}{Prompt for \textsc{Crafter} Native TWI}
\begin{center}
    \includegraphics[width=0.5\textwidth]{prompts/direct_vlm_output/crafter-1.png}
    \includegraphics[width=0.5\textwidth]{prompts/direct_vlm_output/crafter-2.png}
\end{center}
\begin{lstlisting}[style=promptstyle]
Output the 64x64 map of this image, with the help of the reference. Suppose left-top is (0,0), and bottom-right is (63, 63). (row, col)  
  
Format:
  
```json  
{"sand": [(..., ...), ...], "grass": [...], ...}  
```
  
Warning: You cannot use any tools other than vanilla "image cropping" to analyze the image. You are especially banned from using any coding besides the simplest cropping without any further visual cues.

This means, besides basic system commands like `cd`, `mkdir`, `ls`, etc., the only Bash command you can use is `magick` to crop the image. Also, you are completely banned from running `python` or any kind of popular programming language script.
\end{lstlisting}
\end{promptbox}
\begin{promptbox}{Prompt for \textsc{CubeBench} Native TWI}
\begin{center}
    \includegraphics[width=0.5\textwidth]{prompts/direct_vlm_output/cubebench-1.png}
\end{center}
\begin{lstlisting}[style=promptstyle]
You are a Rubik's Cube state reader. The image shows each sticker with a global numeric ID.

Read sticker colors by ID from 1 to 54 and build `state` as this exact ID order.

[Color codes]
Only use: R, G, B, Y, O, W

[Strict output requirements]
* Output only one JSON object
* No explanation, no markdown, no extra text
* state must be exactly 54 characters
* state must be exactly colors of ID 1..54 in order
* state characters must be only R/G/B/Y/O/W

[Allowed output format]
{"state":"XXXXXXXXXXXXXXXXXXXXXXXXXXXXXXXXXXXXXXXXXXXXXXXXXXXXXX"}

Warning: You cannot use any tools other than vanilla "image cropping" to analyze the image. You are especially banned from using any coding besides the simplest cropping without any further visual cues.

This means, besides basic system commands like `cd`, `mkdir`, `ls`, etc., the only Bash command you can use is `magick` to crop the image. Also, you are completely banned from running `python` or any kind of popular programming language script.
\end{lstlisting}
\end{promptbox}



\end{document}